\documentclass[11pt]{article}

\usepackage{emnlp2023}
\usepackage{times}
\usepackage{latexsym}

\usepackage[T1]{fontenc}

\usepackage[utf8]{inputenc}
\usepackage{microtype}

\usepackage{inconsolata}
\usepackage{amsmath,amssymb,amsfonts}
\usepackage{graphicx}
\usepackage{textcomp}
\usepackage{tabularx}
\usepackage{hhline}
\usepackage{booktabs}
\usepackage{multirow}
\usepackage{array}
\usepackage{colortbl}
\usepackage{adjustbox}
\usepackage{hyperref}  
\usepackage{makecell}
\usepackage[noend]{algpseudocode}
\usepackage{algorithm}
\usepackage{bm}
\usepackage{tcolorbox} 
\usepackage{url} 
\definecolor{lav1}{HTML}{FEFEFF}  
\definecolor{lav2}{HTML}{FCF9FF}  
\definecolor{lav3}{HTML}{F8F1FF}  
\definecolor{lav4}{HTML}{F2E8FF}  
\definecolor{lav5}{HTML}{EBDEFF}  
\definecolor{lav6}{HTML}{E2D1FF}  
\definecolor{lav7}{HTML}{D8C2FF}  
\newcommand{\lavcell}[1]{%
    \ifdim #1pt < 0.2pt
        \cellcolor{lav1}#1
    \else\ifdim #1pt < 0.35pt
        \cellcolor{lav2}#1
    \else\ifdim #1pt < 0.45pt
        \cellcolor{lav3}#1
    \else\ifdim #1pt < 0.55pt
        \cellcolor{lav4}#1
    \else\ifdim #1pt < 0.65pt
        \cellcolor{lav5}#1
    \else\ifdim #1pt < 0.75pt
        \cellcolor{lav6}#1
    \else
        \cellcolor{lav7}#1
    \fi\fi\fi\fi\fi\fi
}
\newcommand{\lavmax}[1]{%
    \ifdim #1pt < 0.2pt
        \cellcolor{lav1}\textbf{#1}
    \else\ifdim #1pt < 0.35pt
        \cellcolor{lav2}\textbf{#1}
    \else\ifdim #1pt < 0.45pt
        \cellcolor{lav3}\textbf{#1}
    \else\ifdim #1pt < 0.55pt
        \cellcolor{lav4}\textbf{#1}
    \else\ifdim #1pt < 0.65pt
        \cellcolor{lav5}\textbf{#1}
    \else\ifdim #1pt < 0.75pt
        \cellcolor{lav6}\textbf{#1}
    \else
        \cellcolor{lav7}\textbf{#1}
    \fi\fi\fi\fi\fi\fi
}
\newcommand{\lavcellbig}[1]{%
    \ifdim #1pt < 15pt
        \cellcolor{lav1}#1
    \else\ifdim #1pt < 25pt
        \cellcolor{lav2}#1
    \else\ifdim #1pt < 30pt
        \cellcolor{lav3}#1
    \else\ifdim #1pt < 35pt
        \cellcolor{lav4}#1
    \else\ifdim #1pt < 40pt
        \cellcolor{lav5}#1
    \else\ifdim #1pt < 50pt
        \cellcolor{lav6}#1
    \else
        \cellcolor{lav7}#1
    \fi\fi\fi\fi\fi\fi
}
\newcommand{\lavmaxbig}[1]{%
    \ifdim #1pt < 15pt
        \cellcolor{lav1}\textbf{#1}
    \else\ifdim #1pt < 25pt
        \cellcolor{lav2}\textbf{#1}
    \else\ifdim #1pt < 30pt
        \cellcolor{lav3}\textbf{#1}
    \else\ifdim #1pt < 35pt
        \cellcolor{lav4}\textbf{#1}
    \else\ifdim #1pt < 40pt
        \cellcolor{lav5}\textbf{#1}
    \else\ifdim #1pt < 50pt
        \cellcolor{lav6}\textbf{#1}
    \else
        \cellcolor{lav7}\textbf{#1}
    \fi\fi\fi\fi\fi\fi
}
\newcommand{\lavcellrev}[1]{%
    \ifdim #1pt < 15pt
        \cellcolor{lav7}#1  
    \else\ifdim #1pt < 25pt
        \cellcolor{lav6}#1
    \else\ifdim #1pt < 30pt
        \cellcolor{lav5}#1
    \else\ifdim #1pt < 35pt
        \cellcolor{lav4}#1
    \else\ifdim #1pt < 40pt
        \cellcolor{lav3}#1
    \else\ifdim #1pt < 50pt
        \cellcolor{lav2}#1
    \else
        \cellcolor{lav1}#1
    \fi\fi\fi\fi\fi\fi
}
\newcommand{\lavminrev}[1]{%
    \ifdim #1pt < 15pt
        \cellcolor{lav7}\textbf{#1}  
    \else\ifdim #1pt < 25pt
        \cellcolor{lav6}\textbf{#1}
    \else\ifdim #1pt < 30pt
        \cellcolor{lav5}\textbf{#1}
    \else\ifdim #1pt < 35pt
        \cellcolor{lav4}\textbf{#1}
    \else\ifdim #1pt < 40pt
        \cellcolor{lav3}\textbf{#1}
    \else\ifdim #1pt < 50pt
        \cellcolor{lav2}\textbf{#1}
    \else
        \cellcolor{lav1}\textbf{#1}
    \fi\fi\fi\fi\fi\fi
}
\newcommand{\lavcellrm}[1]{%
    \ifdim #1pt < -1pt
        \cellcolor{lav1}#1
    \else\ifdim #1pt < 0pt
        \cellcolor{lav2}#1
    \else\ifdim #1pt < 1pt
        \cellcolor{lav3}#1
    \else\ifdim #1pt < 2pt
        \cellcolor{lav4}#1
    \else\ifdim #1pt < 3pt
        \cellcolor{lav5}#1
    \else\ifdim #1pt < 4pt
        \cellcolor{lav6}#1
    \else
        \cellcolor{lav7}#1
    \fi\fi\fi\fi\fi\fi
}
\newcommand{\lavmaxrm}[1]{%
    \ifdim #1pt < -1pt
        \cellcolor{lav1}\textbf{#1}
    \else\ifdim #1pt < 0pt
        \cellcolor{lav2}\textbf{#1}
    \else\ifdim #1pt < 1pt
        \cellcolor{lav3}\textbf{#1}
    \else\ifdim #1pt < 2pt
        \cellcolor{lav4}\textbf{#1}
    \else\ifdim #1pt < 3pt
        \cellcolor{lav5}\textbf{#1}
    \else\ifdim #1pt < 4pt
        \cellcolor{lav6}\textbf{#1}
    \else
        \cellcolor{lav7}\textbf{#1}
    \fi\fi\fi\fi\fi\fi
}

\title{UCO: A Multi-Turn Interactive Reinforcement Learning Method for Adaptive Teaching with Large Language Models}

\author{
    \textbf{Shouang Wei}\textsuperscript{1}, 
    \textbf{Min Zhang}\textsuperscript{1,$\ddagger$,$\dagger$}, 
    \textbf{Xin Lin}\textsuperscript{1,$\dagger$},
    \textbf{Bo Jiang}\textsuperscript{1}, 
    \textbf{Kun Kuang}\textsuperscript{2}, \textbf{Jingyuan Chen}\textsuperscript{2}, \textbf{Zhongxiang Dai}\textsuperscript{3} \\
    \textsuperscript{1}East China Normal University,\\
    \textsuperscript{2}Zhejiang University,
    \textsuperscript{3}The Chinese University of Hong Kong, Shenzhen
    \\
    \small \textsuperscript{$\ddagger$}Project leader \quad 
    \href{52285901019@stu.ecnu.edu.cn}{52285901019@stu.ecnu.edu.cn} \quad
    \textsuperscript{$\dagger$}Corresponding authors: \href{mailto:mzhang@cs.ecnu.edu.cn}{mzhang@cs.ecnu.edu.cn}, \href{xlin@cs.ecnu.edu.cn}{xlin@cs.ecnu.edu.cn}
}

\begin{document}
\maketitle

\begin{abstract}
Large language models (LLMs) are shifting from answer providers to intelligent tutors in educational settings. However, current supervised fine-tuning methods only learn surface-level teaching patterns without dynamic adaptation. Recent reinforcement learning approaches address this, yet two key challenges remain. First, they evaluate teaching effectiveness solely based on whether students produce correct outputs, unable to distinguish whether students genuinely understand or echo teacher-provided answers. Second, they cannot perceive students' evolving cognitive states through dialogue, failing to adapt teaching strategies accordingly. We propose Unidirectional Cognitive Optimization (UCO) to address these challenges. UCO employs a multi-turn interactive reinforcement learning paradigm with two synergistic reward functions: the Progress Reward captures cognitive advancement, evaluating students' transition from confusion to comprehension, and the Scaffold Reward dynamically identifies each student's Zone of Proximal Development (ZPD), encouraging the teacher to conduct productive teaching within this zone. We evaluate UCO against 11 baseline models on BigMath and MathTutorBench. Results demonstrate that UCO outperforms all equivalent-scale models and achieves performance comparable to leading closed-source models. Code is available at \url{https://github.com/ECNU-RAIL/UCO-EMNLP2026}.
\end{abstract}

\section{Introduction}
Large language models are shifting from "answer providers" to "intelligent tutors" in educational applications. Mainstream methods rely on supervised fine-tuning (SFT) with human-annotated or model-generated teaching dialogues~\cite{macina2023mathdial,wang2024problem,liu2024socraticlm}, showing how teachers guide students through questioning, prompting, and feedback. However, this imitation approach faces a key challenge: 
\begin{figure}[htbp]
    \centering
    \includegraphics[width=0.42\textwidth]{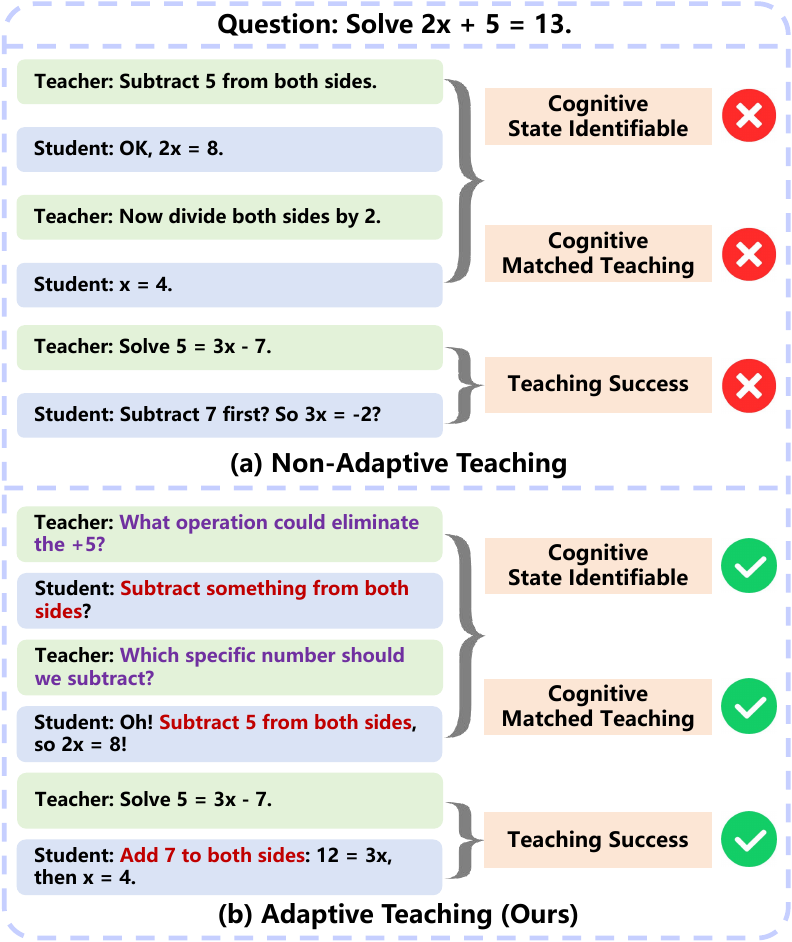}
    \caption{Two Teaching Methods: Comparison Between Non-Adaptive (a) and Adaptive (b) approaches.}
    \label{fig:example}
\end{figure}
SFT-trained models repeat fixed patterns from training data, lacking the dynamic adaptability of real teachers.

Current research explores reinforcement learning (RL) methods to break through this bottleneck~\cite{fahid2024online,riedmann2025reinforcement,dinucu2025problem,qu2026tpop}. Early work primarily adopts single-turn RL paradigms~\cite{zhou2024large,castlejia2026,poesia2024learning,sheng2025learning,hong2026maspob,kong2026workflowr1}. However, real teaching is inherently a multi-turn dynamic process. Thus, recent research shifts to multi-turn RL methods~\cite{dinucu2025problem,xiongbuilding,li2025beyond}, enabling iterative teacher-student interactions. Nevertheless, multi-turn RL approaches primarily rely on the correctness of student outputs as evaluation criteria, leading to two critical deficiencies. First, relying solely on answer correctness as the reward signal, when students output correct answers, models cannot distinguish superficial correctness from genuine understanding (students may merely guess correctly or repeat answers that the teacher revealed), and consequently cannot identify which teaching steps truly promote cognitive improvement. Second, since the current reward function design prevents models from perceiving students' actual understanding levels, models cannot select teaching strategies that match students' comprehension. As illustrated in Figure~\ref{fig:example}(a), this non-adaptive teaching fails to identify student cognitive states and provide cognitive-matched strategies, leading to teaching failure.

To effectively address these limitations, we propose the \textbf{Unidirectional Cognitive Optimization (UCO)} method. UCO adopts a multi-turn interactive reinforcement learning paradigm. The teacher and student models interact continuously to generate online training rollouts dynamically. The approach computes rollout returns based on reward mechanisms and uses GRPO~\cite{shao2024deepseekmath} to update the teacher policy. As shown in Figure~\ref{fig:example}(b), our adaptive teaching approach identifies student cognitive states and provides matched teaching strategies. This method ultimately achieves teaching success. The core innovation lies in two specifically designed reward functions:

\begin{itemize}
\item \textbf{Progress Reward.} From an information theory perspective, we formalize student cognitive progress as an entropy reduction process~\cite{press2024entropy,davis2007information}. Students transition from high-entropy states with high uncertainty to low-entropy states with high certainty. However, computing the entropy of student cognitive states is impractical. We argue that increases in student model probability for outputting correct answers essentially equal cognitive state transitions. Based on this insight, we design dual-dimensional indicators to quantify this entropy reduction process. We combine these two dimensions with weighted aggregation to provide reward signals for each interaction turn.

\begin{enumerate}
    \item \textbf{Potential Capability Score:} For each teacher output, we first use an oracle model to generate multiple candidate correct student responses. We then compute the log-probability of the student model for generating each candidate response. Higher values indicate the student model has high confidence in the correct response. This corresponds to cognitive entropy reduction. Lower values reflect student confusion states. We apply the hyperbolic tangent function to map log-probabilities to the $[-1, 1]$, yielding the Potential Capability Score.
    
    \item \textbf{Semantic Quality Score:} We use text embedding models to compute maximum cosine similarity between student outputs and candidate correct answers. We designate this similarity measure as the Semantic Quality Score. 
\end{enumerate}

\item \textbf{Scaffold Reward.} 
Pedagogical practice reveals that overly simplistic tasks induce boredom, whereas prohibitively challenging ones elicit frustration. Difficulty should remain within the student's "productive struggle" range, termed by Vygotsky as the Zone of Proximal Development (ZPD)~\cite{shabani2010vygotsky}. To translate this into a computable mechanism, we divide teacher behaviors by cognitive load into five ordered levels, from high to low: Metacognitive, Strategic, Conceptual, Step-by-Step, and Example Demonstration. Leveraging this tiered scaffolding framework, we operationalize a mechanism for adaptive ZPD demarcation. We first evaluate the student model's success probability under each scaffold level and identify the highest success level as mastery ability. We then reduce scaffolding by one level, moving tasks into the ZPD region requiring moderate productive struggle. The system provides positive rewards when the selected scaffold falls within the ZPD, and applies proportionally increasing penalties for deviations.
\end{itemize}

By combining these reward functions via weighted aggregation, the method computes rewards after each turn. Then we update the teacher model's policy for cognition-oriented adaptive teaching optimization. Our key contributions are:

\begin{itemize}
\item We propose a novel unidirectional cognitive optimization method that generates training trajectories online through iterative teacher-student interactions, dynamically adjusting teaching strategies as student cognitive states evolve, enabling truly adaptive instruction.

\begin{figure*}[htbp]
    \centering
    \includegraphics[width=1.0\textwidth]{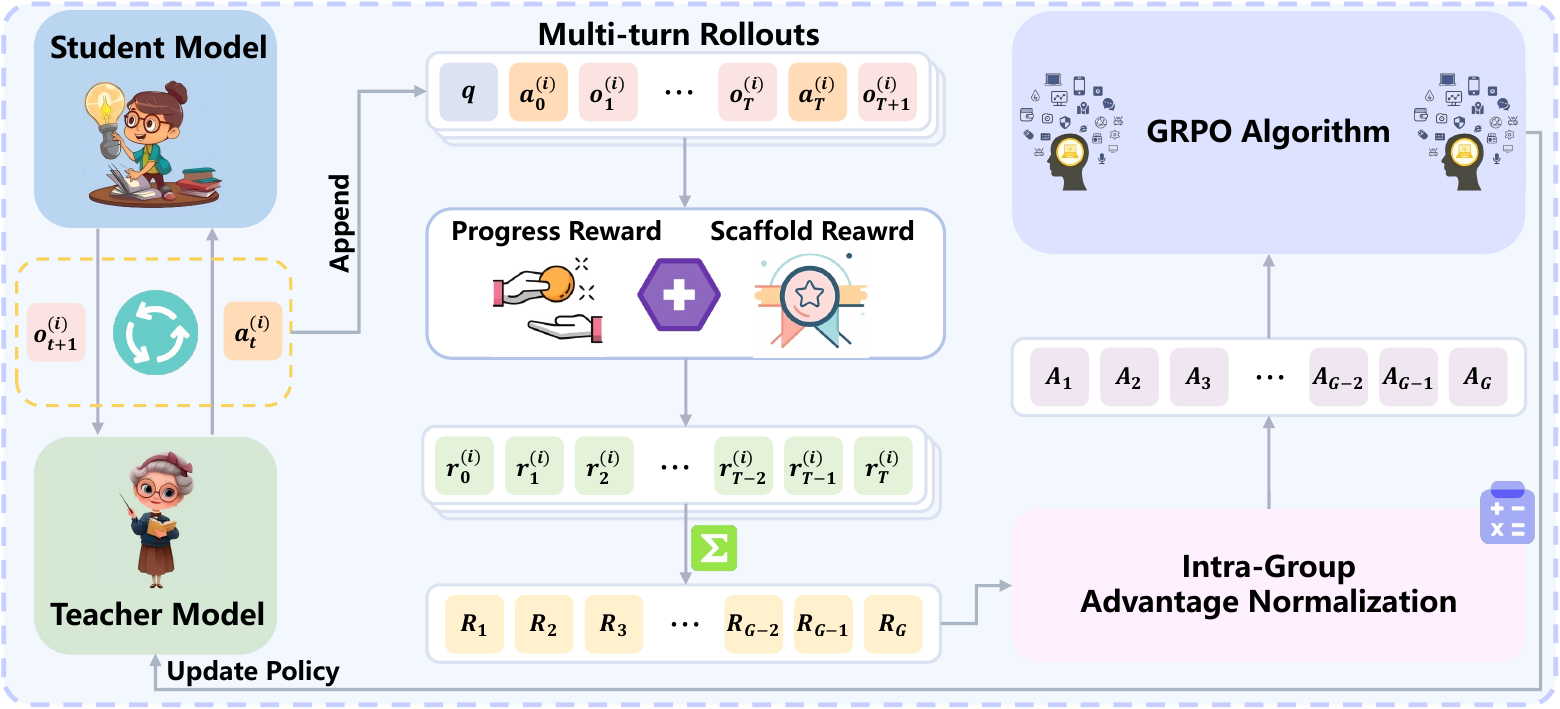}
    \caption{Overview of the Unidirectional Cognitive Optimization (UCO) method, which optimizes the teacher model through multi-turn interactions with the student model, designing progress reward and scaffold reward to evaluate teacher outputs at each turn, and updating the teacher model via GRPO.}
    \label{fig:F0}
\end{figure*}
\item We design a cognition-oriented reward function with two components: a progress reward quantifying cognitive improvement, and a ZPD-based scaffolding reward. These rewards synergistically ensure guidance aligned with students' cognitive levels.

\item We compare our method against 11 baselines on BigMath and MathTutorBench. Our UCO model surpasses all comparable-sized models and achieves performance comparable to advanced closed-source models.
\end{itemize}

\section{Related Work}
\textbf{Math Tutoring with Large Language Models.}
Effective teaching builds student cognition through scaffolding guidance rather than direct answers~\cite{van2010scaffolding,bliss1996effective,rosenshine1992use}.
However, LLMs center their training on answer generation, fundamentally conflicting with pedagogical principles~\cite{tack2022ai,kucheria2025comparing,puech2024towards,macina2023opportunities,codolzhang2025}.
Transforming LLMs into qualified teaching partners becomes an active research direction.
Mathematics serves as an ideal testbed due to its logical rigor and quantifiable nature~\cite{luowizardmath,liu2024mathbench,jia2026diacdm}.
Early research constructs structured educational dialogue data through multi-agent simulation~\cite{liu2024socraticlm,de2024tutorchat}, establishing the foundation for learning teaching skills.
Researchers then build RL-specific benchmarks like Big-Math~\cite{albalak2025big} and MathTutorBench~\cite{macina2025mathtutorbench}, advancing research standardization.
To enable dynamic optimization of teaching strategies, researchers adopt RL as the mainstream framework.
One approach introduces offline preference learning to improve teaching content accuracy~\cite{xiongbuilding,chu2025llm}, optimizing problem-solving trajectory quality.
Yet this offline nature prevents real-time strategy adjustment based on student feedback, lacking dynamic interaction capabilities.
Subsequent research builds online interactive environments with simulated students~\cite{dinucu2025problem}, directly rewarding specific teaching behaviors.
However, they rely on simple behavioral rewards, failing to measure the long-term goal of student understanding.
We design an online RL framework to overcome these limitations.
Our teacher model optimizes by real-time perceiving student cognitive states through interaction, achieving truly adaptive teaching in authentic interactive contexts.

\noindent\textbf{Reward Mechanisms for Teaching Dialogues.}
Reinforcement learning formulates teaching dialogues as sequential decision-making processes~\cite{klissarovmodeling,wang2024language}, optimizing pedagogical strategies through reward signals. The reward function architecture serves as the primary determinant of system performance.
Early work employs outcome-oriented sparse rewards~\cite{peng2025agentic,liao2024mario,zhong2024debug}, providing feedback solely at dialogue termination based on response accuracy. This faces severe credit assignment problems as multi-turn dialogues make identifying critical teaching actions difficult, leading to inefficient policy learning.
Subsequent research shifts to process-oriented dense rewards~\cite{zhang2025lessons,yin2025dynamic,liprocess}, evaluating teaching effectiveness after each dialogue turn. However, these mechanisms suffer from goal alienation by focusing only on student output correctness while ignoring whether students understand the content. This encourages teacher models to take shortcuts, receiving high rewards whenever student models output correct steps, regardless of genuine understanding.
We propose a new reward function to address these limitations by analyzing student model cognitive state changes to evaluate cognitive progress and measuring how teaching interventions match the student's proximal development zone. This achieves a paradigm shift from answer correctness to understanding depth.

\section{Method}

\subsection{Problem Formalization}

We formalize adaptive teaching dialogue generation as a POMDP $\mathcal{M} = (\mathcal{S}, \mathcal{A}, \mathcal{O}, \mathcal{T}, \mathcal{R}, \gamma)$, where $\mathcal{S}$ contains problem descriptions and dialogue history, $\mathcal{A}$ represents pedagogical strategies, $\mathcal{O}$ corresponds to student responses, $\mathcal{T}$ models state transitions, $\mathcal{R}$ evaluates teaching effectiveness, and $\gamma$ balances immediate and future rewards.

As illustrated in Figure~\ref{fig:F0}, we propose Unidirectional Cognitive Optimization (UCO) to train an optimal teacher model $\pi_{\text{teacher}}^*$ that adaptively guides students to solve mathematical problems. 
Specifically, UCO tightly couples three core components into a joint optimization loop: (1) multi-turn interactive rollout generation driven by the teacher and student models, (2) cognitive state evaluation via a dual-objective reward function assisted by an oracle model, and (3) teacher policy updates using GRPO. We aim to maximize the expected cumulative reward across dialogue interactions:

\begin{equation}
\label{eq1}
\begin{split}
\pi_{\text{teacher}}^* = \arg\max_{\pi_{\text{teacher}}} \mathbb{E}_{q \sim \mathcal{D}, \tau \sim \rho} \left[ \sum_{t=0}^{T} \gamma^t r_t^{(i)} \right],
\end{split}
\end{equation}
where $\pi_{\text{student}}^{\text{fixed}}$ is the fixed student model, $q$ is a problem sampled from distribution $\mathcal{D}$, $\tau$ is a dialogue rollout sampled from $\rho = \rho(\pi_{\text{teacher}}, \pi_{\text{student}}^{\text{fixed}})$, which denotes the distribution of interactions between the teacher and student models, and $T$ is the maximum dialogue turns. The state at turn $t$ for the $i$-th rollout is $s_t^{(i)} = \{q, a_0^{(i)}, o_1^{(i)}, \ldots, a_{t-1}^{(i)}, o_t^{(i)}\}$, comprising the problem and dialogue history up to turn $t$, with reward $r_t^{(i)} = \mathcal{R}(s_t^{(i)}, a_t^{(i)})$.

\subsection{Dual-Objective Reward Function}
We design a dual-objective reward function to evaluate teaching effectiveness, comprising the progress reward and the scaffold reward.

\subsubsection{\textbf{Progress Reward}}
Based on Shannon's information theory, effective teaching should maximize information gain, guiding students from uncertainty to certainty. This shift from high entropy to low entropy is termed \textbf{Cognitive Progress}. Formally, given a student's cognitive state $\psi_t$ at time $t$, cognitive progress after action $a_t$ is measured by entropy reduction:
\begin{equation}
    \Delta H_t = H(\psi_t) - H(\psi_{t+1}|a_t),
\end{equation}
where $H(\cdot)$ denotes information entropy.

However, directly calculating $H(\psi_t)$ is impractical. We observe that increases in correct output probability essentially reflect transitions from high-entropy to low-entropy states. Based on this, we design $r_{\text{progress}}$ to approximate cognitive progress through two dimensions: the student's \textbf{potential capacity} and \textbf{semantic quality} of outputs.

\noindent\textbf{Potential Capability Score.}
This term directly measures the confidence of the fixed student model $\pi_S^{\text{fixed}}$ in correct paths. We use it as a core proxy for cognitive progress. We calculate it as:

\begin{equation}
\begin{split}
f_{\text{potential}} = \tanh\Bigg(\alpha \cdot \max_{k=1}^{n} \sum_{j=1}^{m_k} \\
\log \pi_S^{\text{fixed}}(c_{t,j}^k | s_t, a_t, c_{t,1:j-1}^k)\Bigg),
\end{split}
\end{equation}
where $\{c_t^1, c_t^2, ..., c_t^n\}$ is the set of candidate correct responses generated by the oracle model for the student model in the $t$-th interaction round. Each candidate $c_t^k$ is a token sequence of length $m_k$, where $c_{t,j}^k$ denotes the $j$-th token in candidate $k$. The term $\max_{k=1}^{n}$ captures the highest log probability that the student model assigns across all oracle-generated candidates, computed autoregressively over the token sequence. This reflects the student model's peak performance potential at the current step. A higher probability value indicates that the student model has greater confidence in generating correct reasoning paths. This corresponds to a cognitive shift from uncertainty (high entropy) to certainty (low entropy), serving as a key signal for measuring cognitive progress.

We introduce the $\tanh$ function and temperature coefficient $\alpha$ to convert log-likelihoods. The range of log-likelihoods is $(-\infty, 0]$. The $\tanh$ function maps values to the bounded range $[-1, 1]$. This ensures numerical stability and penalizes outputs with insufficient confidence. These outputs still reflect the student's confused state. The temperature coefficient $\alpha > 0$ controls reward sensitivity. Through this design, $f_{\text{potential}}$ quantifies the student model's cognitive progress.

\noindent\textbf{Semantic Quality Score.} 
Besides potential ability assessment, we also need to measure whether the student's actual output is semantically close to the correct answer. We define this component as:
\begin{equation}
   f_{\text{semantic}} = \max_{c_j \in \mathcal{C}} \text{sim}(\mathbf{e}(o_{t+1}), \mathbf{e}(c_j)) - \delta,
\end{equation}
where $\mathbf{e}(\cdot)$ denotes a text embedding model (e.g., BAAI General Embedding~\cite{chen2024bge}) that maps text into a vector space, $\text{sim}(\cdot, \cdot)$ measures cosine similarity, and $\mathcal{C}$ is the set of candidate correct answers. The $\max$ operator captures that semantic alignment with any correct answer indicates concept mastery. The bias term $\delta$ filters out semantically ambiguous or low-quality outputs.

We integrate potential ability and semantic quality through the hyperparameter $\lambda \in [0,1]$, capturing both internal ability improvement and external semantic performance to provide a comprehensive cognitive progress signal:
\begin{equation}
   r_{\text{progress}} = \lambda f_{\text{potential}} + (1-\lambda) f_{\text{semantic}}.
\end{equation}

\subsubsection{\textbf{Scaffold Reward}}
The scaffold reward $r_{\text{scaffold}}$ ensures that teaching difficulty is calibrated to students' current abilities. Tasks that are too easy cause students to lose interest, while tasks that are too hard lead to frustration. Ideal teaching maintains task difficulty within a range where students can succeed through moderate effort, a principle Vygotsky termed the Zone of Proximal Development. We transform this theory into a computable mechanism by discretizing teaching actions into five \textbf{scaffold levels} $\mathcal{L}=\{\ell_0, ..., \ell_4\}$ based on cognitive load from high to low: $\ell_0$ (\textbf{Metacognitive Hints}) uses open-ended questions to guide student thinking without providing substantive hints; $\ell_1$ (\textbf{Strategic Hints}) provides macro-level directions without specific methods; $\ell_2$ (\textbf{Conceptual Hints}) identifies key knowledge points or theorems; $\ell_3$ (\textbf{Step-by-Step Hints}) provides specific operational instructions; and $\ell_4$ (\textbf{Example Demonstration Hints}) offers complete solutions to similar problems.

We design a strategy for dynamically identifying the ZPD according to this hierarchy. In each round, the oracle model generates five hints $\{\text{hint}_{\ell_0}, ..., \text{hint}_{\ell_4}\}$ for the student model. We evaluate the probability that the student generates correct responses under each hint:
\begin{equation}
    P(\ell) = \log \pi_S(c_t | s_t, \text{hint}_\ell),
\end{equation}
where $c_t$ represents a sample from the candidate response set, $s_t$ denotes the current student state, and $\text{hint}_\ell$ is the hint at level $\ell$.

We identify the level that maximizes $P(\ell)$ to determine the student's ability range, then lower the target scaffold level by one grade, shifting teaching difficulty from the comfort zone to the ZPD:
\begin{equation}
    \label{eq:l_zpd}
    \ell_{\text{ZPD}} = \max(0, \arg\max_{\ell \in \mathcal{L}} P(\ell) - 1),
\end{equation}
\noindent where the minus-one operation sets the target one level below, creating a moderate challenge. Based on $\ell_{\text{ZPD}}$, we define the scaffolding reward:
\begin{equation}
\label{eq:r_scaffold}
r_{\text{scaffold}} =
\begin{cases}
\sigma(P(\ell(a_t))) + 0.5, & \ell(a_t) = \ell_{\text{ZPD}} \\
-c|\Delta\ell_t|, & \text{otherwise}
\end{cases}
\end{equation}
\noindent where $\text{index}(\cdot)$ converts levels to integers (0 to 4), $P(\ell(a_t))$ is the log probability of correct response, $\sigma$ is the Sigmoid function, $c > 0$ is a penalty coefficient, and $\Delta\ell_t = \text{index}(\ell(a_t)) - \ell_{\text{ZPD}}$ denotes the level distance. When $\ell(a_t) = \ell_{\text{ZPD}}$, the model receives $0.5$ plus a dynamic reward $\sigma(P(\ell(a_t)))$. Otherwise, it receives a penalty proportional to $|\Delta\ell_t|$.
This mechanism enables the teacher to deliver precisely calibrated scaffolding within the ZPD, mitigating both over- and under-scaffolding. The total reward combines both components:
\begin{equation}
r_t^{(i)} = r_{\text{progress}} + r_{\text{scaffold}},
\end{equation}
\noindent where $r_t^{(i)}$ is the reward at round $t$ in rollout $i$.

\subsection{Policy Optimization via GRPO}
We use GRPO to jointly optimize the teacher model $\pi_{\text{teacher}}$ through a continuous feedback loop among the three UCO components. During each training iteration, the components interact as follows: First, the teacher model interacts with the fixed student model $\pi_{\text{student}}^{\text{fixed}}$ to generate $G$ diverse pedagogical trajectories for each problem $q \in \mathcal{B}$. Second, the oracle model is introduced to evaluate the student's hidden cognitive state turn-by-turn, computing $r_t^{(i)}$ to form dense feedback. 
Finally, these rewards are aggregated into a cumulative return $\mathcal{R}_i = \sum_{t=0}^{T} \gamma^t r_t^{(i)}$ for the $i$-th rollout, where $\gamma \in [0, 1]$ is the discount factor. The intra-group advantage is then computed as $A_i = (\mathcal{R}_i - \mu_q) / (\sigma_q + \epsilon)$, with $\mu_q$ and $\sigma_q$ denoting the mean and standard deviation of returns within the same group.
The advantage $A_i$ guides the policy gradient update, with a KL divergence term preventing deviations from the reference policy $\pi_{\text{teacher}}^{\text{ref}}$. We define the final GRPO objective $\mathcal{J}$ as:
\begin{equation}
\label{eq:grpo_objective}
\begin{split}
\mathcal{J} = & \mathbb{E}_{q \sim \mathcal{B}} \Bigg[ \frac{1}{G} \sum_{i=1}^{G} \Bigg( A_i \cdot \sum_{t=0}^{T} \log \pi_{\text{teacher}}(a_t^{(i)} | s_t^{(i)}) \\
& - \xi \cdot D_{\text{KL}}[\pi_{\text{teacher}}(\cdot|s_t^{(i)}) \| \pi_{\text{teacher}}^{\text{ref}}(\cdot|s_t^{(i)})] \Bigg) \Bigg],
\end{split}
\end{equation}
where $\xi > 0$ is the KL penalty weight.  The updated teacher policy is deployed in the next rollout phase, forming closed-loop joint optimization.

\section{Experiments}

\subsection{Experimental Setup}
\textbf{Dataset.} We train on 2,000 randomly sampled BigMath instances~\cite{albalak2025big}, following PedagogicalRL~\cite{dinucu2025problem}, to both validate our method and reduce RL training cost.

\noindent \textbf{Evaluation.} We evaluate tutoring performance on two complementary benchmarks. On BigMath, we follow the exact evaluation protocol of PedagogicalRL, using the same 500 held-out test questions, the same student model (Meta-Llama-3.1-8B-Instruct-FP8~\cite{grattafiori2024llama}), and the same judge model (Gemma-3-27B-IT~\cite{kamath2025gemma}), replacing only the teacher model with ours. We measure: (1) $\Delta$Solve Rate, quantifying student problem-solving improvement; (2) Leak Solution, detecting explicit solution divulgence via LLM adjudication; and (3) Ped-RM micro/macro scores from MathTutorBench~\cite{macina2025mathtutorbench}, assessing pedagogical quality. On MathTutorBench, we follow the evaluation protocol in the original paper and assess nine tasks across three domains: \textit{Math Expertise} (Problem Solving accuracy, Socratic Questioning BLEU), \textit{Student Understanding} (Solution Correctness F1, Mistake Location F1, Mistake Correction accuracy), and \textit{Pedagogy} (win rates on Scaffolding Generation, Pedagogical IF, and Hard variants in dialogues).

\noindent \textbf{Baselines.} We compare against 11 models: three closed-source systems (GPT-4o~\cite{hurst2024gpt}, LearnLM 1.5 Pro and 2.0 Flash~\cite{team2024learnlm}), four open-source models spanning 7B-671B parameters (Qwen2.5 variants~\cite{team2024qwen2}, DeepSeek-V3~\cite{liu2024deepseek}), and four education-specialized models (Qwen2.5-7B-SFT~\cite{macina2023mathdial}, Qwen2.5-7B-MDPO~\cite{xiongbuilding}, SocraticLM~\cite{liu2024socraticlm}, Qwen2.5-7B-RL~\cite{dinucu2025problem}).

\noindent\textbf{Training Setting.} We use Qwen2.5-7B-Instruct~\cite{team2024qwen2} as $\pi_{\text{teacher}}$, Qwen2.5-14B-Instruct~\cite{team2024qwen2} as $\pi_{\text{student}}^{\text{fixed}}$, and Gemini-2.5-Pro-Exp~\cite{comanici2025gemini} for candidate generation (1 teacher response per scaffolding level, $K=5$ student responses per turn). 
Complete details appear in Appendix~\ref{app:hyperparams}.

\subsection{Main Results}
\textbf{Performance on BigMath Benchmark.}
Table~\ref{tab:results2} shows that existing models struggle to balance three dimensions: teaching effectiveness ($\Delta$ Solve Rate), answer leakage (Leak Solution), and teaching quality (Ped-RM micro/macro).

General models prioritize answers over teaching. Closed-source GPT-4o achieves $33.1\%$ $\Delta$ Solve Rate but has $35.2\%$ Leak Solution and only $1.5/-0.3$ Ped-RM scores. Among open-source models, larger parameters worsen the imbalance. The Qwen2.5 series increases $\Delta$ Solve Rate from $11.3\%$ to $38.7\%$ as parameters grow from 7B to 72B, while Leak Solution rises from $29.3\%$ to $61.0\%$ and Ped-RM stays negative. DeepSeek-V3 achieves the highest $\Delta$ Solve Rate of $39.3\%$ but shows $46.6\%$ Leak Solution with $-1.5/-0.8$ Ped-RM scores, revealing large-scale models fundamentally function as ``solvers'' rather than teachers.

Education-specialized models exhibit a different pattern. SFT and MDPO achieve lower Leak Solution rates but their $\Delta$ Solve Rates drop to $8.9\%$ and $16.4\%$. SocraticLM and Qwen2.5-7B-RL achieve better balance, with Qwen2.5-7B-RL reaching $29.1\%$ $\Delta$ Solve Rate and $15.1\%$ Leak Solution, though its Ped-RM scores of $3.6/3.1$ leave room for improvement. LearnLM 2.0 reaches the best Ped-RM scores of $6.8/6.4$ and lowest Leak Solution of $0.9\%$, but its $\Delta$ Solve Rate is only $4.3\%$, demonstrating the inherent trade-off: strict adherence to pedagogical principles can compromise student learning outcomes.

\begin{table}[htbp]
\centering
\renewcommand{\arraystretch}{1.1}
\scriptsize
\setlength{\tabcolsep}{6pt}
\begin{tabular}{l|c|c|c}
\toprule
\textbf{Models} & 
\makecell{\textbf{$\Delta$Solve} \\ \textbf{Rate (\%)}} & 
\makecell{\textbf{Leak} \\ \textbf{Solution (\%)}} & 
\makecell{\textbf{Ped-RM} \\ \textbf{micro/macro}} \\
& {\tiny $\uparrow$} & {\tiny $\downarrow$} & {\tiny $\uparrow$} \\
\midrule
\multicolumn{4}{c}{\textcolor{violet!70}{\textit{Closed-source}}} \\
GPT-4o & 33.1 & 35.2 & 1.5/-0.3 \\
LearnLM 1.5 Pro & 1.5 & 2.6 & 5.9/5.3 \\
LearnLM 2.0 Flash & 4.3 & \textbf{0.9} & \textbf{6.8/6.4} \\
\midrule
\multicolumn{4}{c}{\textcolor{violet!70}{\textit{Open-source}}} \\
Qwen2.5-7B & 11.3 & 29.3 & -0.2/-0.5 \\
Qwen2.5-14B & 29.3 & 41.9 & -0.6/-1.2 \\
Qwen2.5-72B & 38.7 & 61.0 & 1.8/-0.4 \\
DeepSeek-V3 & \textbf{39.3} & 46.6 & -1.5/-0.8 \\
\midrule
\multicolumn{4}{c}{\textcolor{violet!70}{\textit{Education-specific}}} \\
Qwen2.5-7B-SFT & 8.9 & 36.0 & -0.3/-0.7 \\
Qwen2.5-7B-MDPO & 16.4 & 35.6 & 0.2/-0.3 \\
SocraticLM & 15.9 & 40.4 & 1.7/1.7 \\
Qwen2.5-7B-RL & 29.1 & 15.1 & 3.6/3.1 \\
\midrule
\textbf{UCO} & 30.2 & 12.9 & 4.6/4.5 \\
\bottomrule
\end{tabular}
\caption{Performance on BigMath benchmark. Results for all models except UCO are from the PedagogicalRL.}
\label{tab:results2}
\end{table}

UCO achieves synergistic optimization with only 7B parameters. It reaches $30.2\%$ $\Delta$ Solve Rate, surpassing similar-sized models and approaching larger models. UCO achieves $12.9\%$ Leak Solution, representing a $14.6$ percentage point improvement over the strongest teaching baseline Qwen2.5-7B-RL, and $4.6/4.5$ Ped-RM scores ($28\%$ improvement). This three-way balance validates our dual-objective reward mechanism.

Our core advantage lies in modeling student cognitive states. Learning progress reward quantifies cognitive changes, providing rewards only when students truly understand reasoning steps, thus improving $\Delta$ Solve Rate. Scaffolding reward operationalizes ZPD theory by estimating student ability across five levels, identifying the zone of proximal development, and encouraging adaptive strategies that avoid direct answers while preventing inefficient over-questioning. UCO leverages these differentiable cognitive signals, surpassing similarly sized baselines across all three dimensions and achieving comprehensive performance that rivals closed-source models.

\begin{table*}[htbp]
\centering
\renewcommand{\arraystretch}{1.1}
\scriptsize
\setlength{\tabcolsep}{9pt}
\begin{tabular}{l|cc|ccc|cccc}
\toprule
& \multicolumn{2}{c|}{\textbf{Math Expertise}} & 
\multicolumn{3}{c|}{\textbf{Student Understanding}} & 
\multicolumn{4}{c}{\textbf{Pedagogy}} \\
\cmidrule(lr){2-3} \cmidrule(lr){4-6} \cmidrule(lr){7-10}
\multirow{3}{*}{\textbf{Model}} & Problem & Socratic & Solution & Mistake & Mistake & \multicolumn{4}{c}{Teacher response generation} \\
& solving & questioning & correctness & location & correction & scaff. & ped.IF & scaff. & ped.IF \\
& & & & & & & & [hard] & [hard] \\
\cmidrule(lr){2-3} \cmidrule(lr){4-6} \cmidrule(lr){7-10}
& acc. & BLEU & F1 & micro F1 & acc. & \multicolumn{4}{c}{win rate over human teacher} \\
\midrule
\multicolumn{10}{c}{\textcolor{violet!70}{\textit{Education-specific Models}}} \\
Qwen2.5-7B-SFT & 0.77 & 0.24 & 0.27 & \textbf{0.45} & 0.10 & 0.64 & 0.58 & 0.57 & 0.59 \\
Qwen2.5-7B-MDPO & 0.86 & 0.23 & 0.62 & 0.39 & 0.03 & 0.37 & 0.60 & 0.47 & 0.56 \\
SocraticLM & 0.73 & \textbf{0.32} & 0.05 & 0.39 & \textbf{0.23} & 0.39 & 0.39 & 0.28 & 0.28 \\
Qwen2.5-7B-RL & 0.83 & 0.23 & \textbf{0.67} & 0.35 & 0.05 & 0.57 & 0.72 & 0.61 & 0.69 \\
\midrule
\textbf{UCO} & \textbf{0.89} & 0.24 & 0.65 & \textbf{0.45} & 0.08 & \textbf{0.68} & \textbf{0.74} & \textbf{0.72} & \textbf{0.77} \\
\bottomrule
\end{tabular}
\caption{Evaluation on MathTutorBench across math expertise, student understanding, and pedagogy.}
\label{tab:results}
\end{table*}

\begin{table*}[htbp]
\centering
\renewcommand{\arraystretch}{1.1}
\scriptsize
\setlength{\tabcolsep}{8pt}
\begin{tabular}{l|cc|ccc|cccc}
\toprule
& \multicolumn{2}{c|}{\textbf{Math Expertise}} & 
\multicolumn{3}{c|}{\textbf{Student Understanding}} & 
\multicolumn{4}{c}{\textbf{Pedagogy}} \\
\cmidrule(lr){2-3} \cmidrule(lr){4-6} \cmidrule(lr){7-10}
\multirow{3}{*}{\textbf{Model}} & Problem & Socratic & Solution & Mistake & Mistake & \multicolumn{4}{c}{Teacher response generation} \\
& solving & questioning & correctness & location & correction & scaff. & ped.IF & scaff. & ped.IF \\
& & & & & & & & [hard] & [hard] \\
\cmidrule(lr){2-3} \cmidrule(lr){4-6} \cmidrule(lr){7-10}
& acc. & BLEU & F1 & micro F1 & acc. & \multicolumn{4}{c}{win rate over human teacher} \\
\midrule
\multicolumn{10}{c}{\textcolor{violet!70}{\textit{Backbone Model}}} \\
Qwen2.5-7B-Instruct & 0.87 & 0.23 & 0.63 & 0.39 & 0.04 & 0.37 & 0.60 & 0.45 & 0.56 \\
\midrule
\textbf{UCO (full)} & \textbf{0.89} & \textbf{0.24} & \textbf{0.65} & \textbf{0.45} & \textbf{0.08} & \textbf{0.68} & \textbf{0.74} & \textbf{0.72} & \textbf{0.77} \\
\textbf{UCO w/o scaffold reward} & 0.86 & 0.23 & 0.61 & 0.40 & 0.05 & 0.56 & 0.70 & 0.67 & 0.71 \\
\textbf{UCO w/o progress reward} & 0.88 & \textbf{0.24} & 0.63 & 0.42 & 0.07 & 0.61 & 0.71 & 0.69 & 0.74 \\
\bottomrule
\end{tabular}
\caption{Ablation study on MathTutorBench showing impact of reward components across evaluation dimensions.}
\label{tab:ablation_results}
\end{table*}

\noindent\textbf{Comprehensive Evaluation on MathTutorBench.} 
Table~\ref{tab:results} shows evaluation results on MathTutorBench. We compare UCO with education-specific models across nine tasks in three categories: Math Expertise, Student Understanding, and Pedagogy.
Qwen2.5-7B-SFT performs well in Math Expertise but lags significantly in Student Understanding and Pedagogy. Qwen2.5-7B-MDPO improves math solving and student understanding through direct preference optimization, but causes severe Pedagogy degradation. SocraticLM excels in Pedagogy but shows weaknesses in Math Expertise and Student Understanding. Qwen2.5-7B-RL achieves balance across dimensions but lacks overall strength.
In contrast, UCO performs strongly in Math Expertise and Student Understanding while achieving optimal Pedagogy performance. This stems from UCO's cognitive state modeling mechanism, where learning progress rewards track student cognitive changes and scaffolding rewards dynamically adjust teaching strategies, maintaining high-quality teaching in complex multi-turn interactions.

\begin{table}[htbp]
\centering
\renewcommand{\arraystretch}{1.1}
\scriptsize
\setlength{\tabcolsep}{4pt}
\begin{tabular}{l|c|c|c}
\toprule
\textbf{Models} & 
\makecell{\textbf{$\Delta$Solve} \\ \textbf{Rate (\%)}} & 
\makecell{\textbf{Leak} \\ \textbf{Solution (\%)}} & 
\makecell{\textbf{Ped-RM} \\ \textbf{micro/macro}} \\
& {\tiny $\uparrow$} & {\tiny $\downarrow$} & {\tiny $\uparrow$} \\
\midrule
\multicolumn{4}{c}{\textcolor{violet!70}{\textit{Backbone Model}}} \\
Qwen2.5-7B-Instruct & 11.3 & 29.3 & -0.2/-0.5 \\
\midrule
\textbf{UCO (full)} & \textbf{30.2} & \textbf{12.9} & \textbf{4.6}/\textbf{4.5} \\
\textbf{UCO w/o scaffold reward} & 23.5 & 20.2 & 2.2/2.3 \\
\textbf{UCO w/o progress reward} & 28.6 & 15.6 & 4.1/3.9 \\
\bottomrule
\end{tabular}
\caption{Ablation study on BigMath on the contribution of progress and scaffolding reward components.}
\label{tab:ablation}
\end{table}

\subsection{Ablation Study}
\textbf{Ablation Study on BigMath Dataset.}
Table~\ref{tab:ablation} validates the necessity of our reward components through ablation studies on BigMath.
The entire UCO model achieves $30.2\%$ $\Delta$ Solve Rate, $12.9\%$ Leak Solution, and 4.6/4.5 Ped-RM micro/macro.
Removing scaffolding rewards causes the most severe degradation. Compared to the entire setup, $\Delta$ Solve Rate decreases by 6.7 percentage points to $23.5\%$, Leak Solution rises by 7.3 percentage points to $20.2\%$, and Ped-RM micro/macro drops to 2.2/2.3. Without scaffolding guidance, the model fails to adapt to students' zones of proximal development, resulting in inefficient teaching.
Removing progress rewards leads to smaller but notable declines. Compared to the entire setup, $\Delta$ Solve Rate drops by 1.6 percentage points to $28.6\%$, Leak Solution increases by 2.7 percentage points to $15.6\%$, and Ped-RM micro/macro decreases to 4.1/3.9. The model struggles to track student understanding and cannot adjust teaching strategies based on learning states.
All reward configurations outperform the backbone, confirming the effectiveness of our reward design.

\noindent\textbf{Ablation Study on MathTutorBench Dataset.} Table~\ref{tab:ablation_results} further validates our reward function through ablation studies on MathTutorBench.
The entire UCO model performs well across all three dimensions. We remove scaffolding rewards and observe significant declines: $3.4\%$ in Math Expertise, $9.0\%$ in Student Understanding, and $10.6\%$ in Pedagogy. Without scaffolding guidance, the model fails to break down content into progressive steps. It causes a mismatch between teaching strategies and student capabilities.
Removing progress rewards has a greater impact on recognizing student understanding states (Student Understanding) and dynamically adjusting teaching strategies (Pedagogy), but less impact on accurately delivering mathematical knowledge (Math Expertise).
It indicates that progress rewards primarily regulate teaching pace based on student feedback.


\noindent\textbf{Additional Ablation Studies.}
To thoroughly validate UCO, Appendix~\ref{app:ablation} details extensive analyses along four aspects. First, we compare our dense reward mechanism against the sparse PedagogicalRL reward, confirming the superiority of turn-level cognitive feedback (Appendix~\ref{app:reward_mechanism}). We then demonstrate robustness by varying backbone architectures, embedding models, and oracle choices, all yielding consistent improvements (Appendix~\ref{app:backbone}, \ref{app:embedding}, and \ref{app:oracle}). Next, we evaluate the sensitivity of key hyperparameters, including $\lambda$, $\alpha$, $\delta$, $c$, and the number of rollouts, confirming that our defaults optimally balance cognitive scaffolding with leakage prevention (Appendix~\ref{app:rollout}, \ref{app:lambda}, \ref{app:alpha}, \ref{app:delta}, and \ref{app:deviation}). Finally, we provide broader evaluations covering general mathematical reasoning preservation (Appendix~\ref{app:general_math}), human evaluations (Appendix~\ref{app:human_eval}), qualitative case studies against GPT-4o, efficiency analysis, and additional results on MathDial (Appendix~\ref{app:case_study}, ~\ref{app:efficiency}, and \ref{app:mathdial}).


\section{Conclusion}
In this paper, we propose UCO, a cognition-oriented multi-turn reinforcement learning method for adaptive teaching. To address limitations in existing methods that lack true adaptive capability, we model the teaching process as continuous teacher-student interactions and design two synergistic reward functions. The Progress Reward quantifies cognitive progress through entropy reduction, evaluating whether students transition from uncertainty to understanding. The Scaffold Reward dynamically identifies each student's Zone of Proximal Development and encourages teaching strategies that maintain productive struggle within this zone. Extensive experiments on BigMath and MathTutorBench benchmarks demonstrate our UCO model surpasses all equivalent-scale models and achieves performance comparable to advanced closed-source models.
We argue that modeling the teaching process from the perspective of cognitive state evolution applies not only to mathematical tutoring but also to other educational domains such as programming education and scientific reasoning. In future work, we will explore extending UCO to broader educational scenarios.

\section*{Limitations}
We conducted a comprehensive evaluation of the teaching effectiveness of the UCO method, but it has the following limitations.

\begin{enumerate}
    \item We focus on K-12 math tutoring because math problems provide clear answers and verifiable reasoning, enabling precise signal design for reinforcement learning rewards. While the teaching modeling principle based on cognitive state evolution is universally applicable, extending it to other domains faces challenges in reward design and benchmark construction. Future work should develop domain-specific cognitive progress metrics and scaffolding hierarchies for different subjects, and collaborate with researchers in education and psychology to establish evaluation benchmarks.

    \item Our study employs text-based dialogue interaction, aligning with mainstream applications of large language models. However, teaching scenarios involve multimodal information such as graphics, handwritten calculations, and vocal intonation. Future work should extend the UCO framework to multimodal teaching scenarios to leverage such information.
    
    \item The effectiveness of UCO has been validated mainly in offline datasets, and it has not yet been tested in real classrooms. The teaching impact of UCO requires further verification in real-world teaching scenarios.
 
\end{enumerate}

\section*{Acknowledgments}
This study was funded by Guangxi Science and Technology Program (2025AB25069309), the Open Research Fund of Key Laboratory of Advanced Theory and Application in Statistics and Data Science (East China Normal University). This work was supported in part by "Pioneer" and "Leading Goose" R\&D Program of Zhejiang (2025C02037) and the Key R\&D Program of Hangzhou (2025SZDA0254). This work was supported in part by the National Natural Science Foundation of China (Grant No.~62506319).

\bibliography{anthology,custom}
\bibliographystyle{acl_natbib}

\clearpage
\appendix


\newpage
    
    
    

\section{The Use of Large Language Models}
In this paper, we employ LLMs as an auxiliary tool. We utilized LLMs for language polishing to
enhance clarity, and we manually reviewed all modifications.

\section{Implementation Details}
\label{app:hyperparams}

\subsection{Training Configuration}
We conduct all experiments on a server equipped with 8 NVIDIA H200 GPUs. The experimental environment uses Python 3.11, CUDA 12.8, and PyTorch 2.7. Training takes approximately 112 hours for the complete process.

\subsection{Model Architecture}
Table~\ref{tab:model_arch} summarizes the model configurations.

\begin{table}[h]
\centering
\small
\begin{tabular}{@{}lp{5cm}@{}}
\toprule
\textbf{Component} & \textbf{Model \& Purpose} \\
\midrule
Teacher & Qwen2.5-7B-Instruct for $\pi_{\text{teacher}}$ \\
Student & Qwen2.5-14B-Instruct for $\pi_{\text{student}}^{\text{fixed}}$ \\
Generator & Gemini-2.5-Pro for candidate sampling \\
\bottomrule
\end{tabular}
\caption{Model Architecture Configuration}
\label{tab:model_arch}
\end{table}

\subsection{Hyperparameters}
All hyperparameters are determined through grid search, as shown in Table~\ref{tab:hyperparams}.

\begin{table}[h]
\centering
\small
\begin{tabular}{@{}p{2.2cm}p{2.8cm}r@{}}
\toprule
\textbf{Category} & \textbf{Parameter} & \textbf{Value} \\
\midrule
\multirow{8}{*}{\makecell[l]{GRPO\\Optimization}} 
& Learning rate $\eta$ & $5 \times 10^{-6}$ \\
& Batch size $|\mathcal{B}|$ & 16 \\
& Rollouts $G$ & 4 \\
& Max length $T$ & 10 \\
& Discount $\gamma$ & 0.95 \\
& KL weight $\xi$ & 0.01 \\
& Norm. $\epsilon$ & $10^{-8}$ \\
& Optimizer & AdamW \\
\midrule
\multirow{4}{*}{\makecell[l]{Progress\\Reward}} 
& Balance $\lambda$ & 0.5 \\
& Temperature $\alpha$ & 2.0 \\
& Bias $\delta$ & 0.7 \\
& Embedding & BGE \\
\midrule
Scaffold & Penalty $c$ & 0.2 \\
\bottomrule
\end{tabular}
\caption{Hyperparameter Configuration}
\label{tab:hyperparams}
\end{table}

\subsection{Baseline Models}
Table~\ref{tab:baselines} lists all baseline models in our experiments.

\begin{table}[h]
\centering
\small
\begin{tabular}{@{}p{1cm}p{4cm}r@{}}
\toprule
\textbf{Category} & \textbf{Model} & \textbf{Size} \\
\midrule
\multirow{3}{*}{Closed} 
& GPT-4o~\cite{hurst2024gpt} & - \\
& LearnLM 1.5 Pro~\cite{team2024learnlm} & - \\
& LearnLM 2.0 Flash~\cite{team2024learnlm} & - \\
\midrule
\multirow{4}{*}{Open} 
& Qwen2.5-Instruct~\cite{team2024qwen2} & 7B \\
& Qwen2.5-Instruct~\cite{team2024qwen2} & 14B \\
& Qwen2.5-Instruct~\cite{team2024qwen2} & 72B \\
& DeepSeek-V3~\cite{liu2024deepseek} & 671B \\
\midrule
\multirow{4}{*}{Education} 
& Qwen2.5-SFT~\cite{macina2023mathdial} & 7B \\
& Qwen2.5-MDPO~\cite{xiongbuilding} & 7B \\
& SocraticLM~\cite{liu2024socraticlm} & 7B \\
& Qwen2.5-RL~\cite{dinucu2025problem} & 7B \\
\bottomrule
\end{tabular}
\caption{Baseline Model Specifications}
\label{tab:baselines}
\end{table}

\subsection{Evaluation Metrics}
Tables~\ref{tab:bigmath_metrics} and~\ref{tab:mathtutorbench_metrics} show the evaluation metrics.

\begin{table}[h]
\centering
\small
\begin{tabular}{@{}p{2cm}p{4.5cm}@{}}
\toprule
\textbf{Metric} & \textbf{Description} \\
\midrule
$\Delta$Solve & Student improvement before/after dialogue \\
Leak Solution & LLM-detected solution divulgence \\
Ped-RM & Pedagogical quality scores \\
\bottomrule
\end{tabular}
\caption{BigMath Test Set Metrics}
\label{tab:bigmath_metrics}
\end{table}

\begin{table}[h]
\centering
\small
\begin{tabular}{@{}p{2cm}p{2.5cm}p{2cm}@{}}
\toprule
\textbf{Domain} & \textbf{Task} & \textbf{Metric} \\
\midrule
\multirow{2}{*}{\makecell[l]{Math\\Expertise}} 
& Problem Solving & Answer accuracy \\
& Socratic Quest. & BLEU score \\
\midrule
\multirow{3}{*}{\makecell[l]{Student\\Understand.}} 
& Sol. Correctness & F1 micro \\
& Mistake Location & F1 score \\
& Mistake Correct. & Accuracy \\
\midrule
\multirow{3}{*}{Pedagogy} 
& Scaffold Gen. & Win rate \\
& Pedagogical IF & Win rate \\
& Hard versions & Win rate \\
\bottomrule
\end{tabular}
\caption{MathTutorBench Evaluation Tasks}
\label{tab:mathtutorbench_metrics}
\end{table}

\section{Additional Ablation Studies}
\label{app:ablation}

\subsection{\textbf{Impact of Reward Mechanisms on Training Efficiency.}} \label{app:reward_mechanism}
Table~\ref{tab:controlled_comparison} presents a comparison on the BigMath dataset to validate the superiority of our reward design. We ensure a strictly fair comparison by maintaining an identical training pipeline (same GRPO algorithm, 2,000 samples, and 4 rollouts) and only substituting our mechanism with the PedagogicalRL reward. Results show that PedagogicalRL underperforms UCO across all dimensions. Despite its explicit LLM-based hard penalties, PedagogicalRL exhibits higher Leak Solution rates, while its $\Delta$ Solve Rate and teaching quality scores (Ped-RM) lag substantially behind our method.

This performance discrepancy highlights the fundamental difference between sparse and dense reward designs. Specifically, PedagogicalRL relies on sparse, conversation-level rewards tied to the correctness of the final answer, coupled with all-or-nothing penalties for pedagogical violations. Such sparse feedback provides insufficient gradient signals, making it difficult to guide the model toward the optimal policy equilibrium without leaking answers. In contrast, UCO mitigates this by providing dense, turn-level feedback: the Progress Reward evaluates the reduction in entropy of cognitive states, and the Scaffold Reward precisely demarcates the Zone of Proximal Development (ZPD) at each step. By supplying these fine-grained, cognition-oriented signals, UCO achieves this pedagogical equilibrium, mastering adaptive teaching while strictly preventing answer leakage.

\begin{table*}[htbp]
\centering
\renewcommand{\arraystretch}{1.1}
\scriptsize
\setlength{\tabcolsep}{3.5pt}
\begin{tabular}{l|cc|ccc|cccc}
\toprule
& \multicolumn{2}{c|}{\textbf{Math Expertise}} & \multicolumn{3}{c|}{\textbf{Student Understanding}} & \multicolumn{4}{c}{\textbf{Pedagogy}} \\
\cmidrule(lr){2-3} \cmidrule(lr){4-6} \cmidrule(lr){7-10}
\textbf{Model} & 
\makecell{Problem \\ solving \\ acc.} & 
\makecell{Socratic \\ questioning \\ BLEU} & 
\makecell{Solution \\ correctness \\ F1} & 
\makecell{Mistake \\ location \\ micro F1} & 
\makecell{Mistake \\ correction \\ acc.} & 
\makecell{scaff. \\ win rate} & 
\makecell{ped.IF \\ win rate} & 
\makecell{scaff. [hard] \\ win rate} & 
\makecell{ped.IF [hard] \\ win rate} \\
\midrule
\multicolumn{10}{c}{\textcolor{violet!70}{\textit{LLaMA3.2-3B-Instruct as Backbone Model}}} \\
\midrule
LLaMA3.2-3B-Instruct & 0.60 & 0.29 & 0.67 & 0.41 & 0.13 & 0.64 & 0.63 & 0.45 & 0.40 \\
\textbf{LLaMA3.2-3B-Instruct + UCO} & \textbf{0.63} & \textbf{0.30} & \textbf{0.69} & \textbf{0.44} & \textbf{0.16} & \textbf{0.69} & \textbf{0.68} & \textbf{0.51} & \textbf{0.46} \\
\midrule
\multicolumn{10}{c}{\textcolor{violet!70}{\textit{LLaMA3.1-8B-Instruct as Backbone Model}}} \\
\midrule
LLaMA3.1-8B-Instruct & 0.70 & 0.29 & 0.63 & 0.29 & 0.09 & 0.61 & 0.67 & 0.46 & 0.49 \\
\textbf{LLaMA3.1-8B-Instruct + UCO} & \textbf{0.75} & \textbf{0.30} & \textbf{0.67} & \textbf{0.34} & \textbf{0.13} & \textbf{0.67} & \textbf{0.72} & \textbf{0.53} & \textbf{0.55} \\
\bottomrule
\end{tabular}
\caption{Cross-architecture generalizability on MathTutorBench using 500 training samples.}
\label{tab:llama_generalizability}
\end{table*}

\begin{table}[htbp]
\centering
\renewcommand{\arraystretch}{1.1}
\scriptsize
\setlength{\tabcolsep}{5.5pt}
\begin{tabular}{l|c|c|c}
\toprule
\textbf{Reward Function} & 
\makecell{\textbf{$\Delta$Solve} \\ \textbf{Rate (\%)}} & 
\makecell{\textbf{Leak} \\ \textbf{Solution (\%)}} & 
\makecell{\textbf{Ped-RM} \\ \textbf{micro/macro}} \\
& {\tiny $\uparrow$} & {\tiny $\downarrow$} & {\tiny $\uparrow$} \\
\midrule
\multicolumn{4}{c}{\textcolor{violet!70}{\textit{Backbone Model}}} \\
Qwen2.5-7B-Instruct & 11.3 & 29.3 & -0.2/-0.5 \\
\midrule
\multicolumn{4}{c}{\textcolor{violet!70}
} \\
PedagogicalRL Reward & 24.3 & 19.2 & 2.8/2.7 \\
\textbf{UCO Reward (Ours)} & \textbf{30.2} & \textbf{12.9} & \textbf{4.6}/\textbf{4.5} \\
\bottomrule
\end{tabular}
\caption{Comparison between PedagogicalRL and UCO reward on BigMath.}
\label{tab:controlled_comparison}
\end{table}

\subsection{\textbf{Impact of Backbone Model.}} \label{app:backbone}
To evaluate the generalization capability of UCO across various model architectures, we conducted experiments using LLaMA3.1-8B-Instruct~\cite{grattafiori2024llama} and LLaMA3.2-3B-Instruct~\cite{grattafiori2024llama} as backbone models. While keeping all other experimental settings constant, we validated the performance on MathTutorBench using 500 training samples.
As shown in Table \ref{tab:llama_generalizability}, UCO consistently improves the performance of backbone models across different architectures and parameter scales. This transferability stems from the design logic of UCO reward functions: both \textit{Progress Reward} and \textit{Scaffold Reward} quantify cognitive states by analyzing the output probability distributions of the student model, rather than relying on architecture-specific features. Consequently, these feedback signals, grounded in the evolution of cognitive states, exhibit high universality, enabling UCO to generalize effectively across models of diverse scales and structures.

\subsection{\textbf{Impact of Embedding Models.}}\label{app:embedding}
To verify UCO's sensitivity to the choice of embedding models, we replace the BGE-M3 with two architecturally distinct models of comparable parameter scale: Multilingual-E5-Large-Instruct \textbf{(Multilingual-E5)}~\cite{wang2024multilingual} and Qwen3-Embedding-0.6B \textbf{(Qwen3-Emb-0.6B)}~\cite{zhang2025qwen3}. As shown in Table~\ref{tab:emb_pedagogy} and Table~\ref{tab:emb_math}, no single embedding model achieves comprehensive superiority across all evaluation dimensions. Regardless of the embedding model employed, the overall performance of UCO substantially surpasses that of the original base model. This result compellingly confirms that UCO is robust to the selection of embedding models.

\begin{table*}[htbp]
\centering
\renewcommand{\arraystretch}{1.1}
\scriptsize
\setlength{\tabcolsep}{3.5pt}
\begin{tabular}{l|cc|ccc|cccc}
\toprule
& \multicolumn{2}{c|}{\textbf{Math Expertise}} & \multicolumn{3}{c|}{\textbf{Student Understanding}} & \multicolumn{4}{c}{\textbf{Pedagogy}} \\
\cmidrule(lr){2-3} \cmidrule(lr){4-6} \cmidrule(lr){7-10}
\textbf{Model} & 
\makecell{Problem \\ solving \\ acc.} & 
\makecell{Socratic \\ questioning \\ BLEU} & 
\makecell{Solution \\ correctness \\ F1} & 
\makecell{Mistake \\ location \\ micro F1} & 
\makecell{Mistake \\ correction \\ acc.} & 
\makecell{scaff. \\ win rate} & 
\makecell{ped.IF \\ win rate} & 
\makecell{scaff. [hard] \\ win rate} & 
\makecell{ped.IF [hard] \\ win rate} \\
\midrule
\multicolumn{10}{c}{\textcolor{violet!70}{\textit{Backbone Model}}} \\
Qwen2.5-7B-Instruct & 0.87 & 0.23 & 0.63 & 0.39 & 0.04 & 0.37 & 0.60 & 0.45 & 0.56 \\
\midrule
UCO w/ BGE-M3 (default) & \textbf{0.89} & \textbf{0.24} & \textbf{0.65} & \textbf{0.40} & \textbf{0.06} & 0.60 & 0.66 & 0.63 & 0.68 \\
UCO w/ multilingual-e5 & 0.88 & 0.23 & 0.64 & \textbf{0.40} & \textbf{0.06} & 0.59 & \textbf{0.67} & \textbf{0.64} & \textbf{0.70} \\
UCO w/ Qwen3-Emb-0.6B & \textbf{0.89} & \textbf{0.24} & \textbf{0.65} & 0.39 & 0.05 & \textbf{0.62} & 0.66 & 0.63 & 0.69 \\
\bottomrule
\end{tabular}
\caption{Performance on MathTutorBench across different embedding models (500 training samples).}
\label{tab:emb_math}
\end{table*}

\begin{table}[htbp]
\centering
\renewcommand{\arraystretch}{1.1}
\scriptsize
\setlength{\tabcolsep}{0.6pt}
\begin{tabular}{lccc}
\toprule
\textbf{Model} & \makecell{\textbf{$\Delta$Solve Rate} \\ \textbf{(\%) $\uparrow$}} & \makecell{\textbf{Leak Solution} \\ \textbf{(\%) $\downarrow$}} & \makecell{\textbf{Ped-RM} \\ \textbf{micro/macro $\uparrow$}} \\
\midrule
\multicolumn{4}{c}{\textcolor{violet!70}{\textit{Backbone Model}}} \\
Qwen2.5-7B-Instruct & 11.3 & 29.3 & -0.2 / -0.5 \\
\midrule
UCO w/ BGE-M3 (default) & 22.8 & \textbf{16.4} & 3.6 / 3.5 \\
UCO w/ Multilingual-E5 & 22.5 & 16.7 & \textbf{3.7} / \textbf{3.7} \\
UCO w/ Qwen3-Emb-0.6B & \textbf{23.2} & 16.9 & 3.5 / 3.4 \\
\bottomrule
\end{tabular}
\caption{Performance on BigMath across different embedding models (500 training samples).}
\label{tab:emb_pedagogy}
\end{table}

\begin{table}[htbp]
\centering
\renewcommand{\arraystretch}{1.1}
\scriptsize
\setlength{\tabcolsep}{0.05pt}
\begin{tabular}{lccc}
\toprule
\textbf{Model} & \makecell{\textbf{$\Delta$Solve Rate} \\ \textbf{(\%) $\uparrow$}} & \makecell{\textbf{Leak Solution} \\ \textbf{(\%) $\downarrow$}} & \makecell{\textbf{Ped-RM} \\ \textbf{micro/macro $\uparrow$}} \\
\midrule
\multicolumn{4}{c}{\textcolor{violet!70}{\textit{Backbone Model}}} \\
Qwen2.5-7B-Instruct & 11.3 & 29.3 & -0.2 / -0.5 \\
\midrule
UCO w/ Gemini-2.5-Pro & 22.8 & \textbf{16.4} & \textbf{3.6 / 3.5} \\
UCO w/ DeepSeek-V3.2 & \textbf{23.1} & 17.3 & 3.2 / 3.1 \\
UCO w/ Qwen3-235B-A22B & 22.6 & 17.6 & 3.0 / 2.9 \\
\bottomrule
\end{tabular}
\caption{Comprehensive evaluation on BigMath across various oracle models (500 training samples).}
\label{tab:oracle_pedagogy}
\end{table}

\begin{table*}[htbp]
\centering
\renewcommand{\arraystretch}{1.1}
\scriptsize
\setlength{\tabcolsep}{3.5pt}
\begin{tabular}{l|cc|ccc|cccc}
\toprule
& \multicolumn{2}{c|}{\textbf{Math Expertise}} & \multicolumn{3}{c|}{\textbf{Student Understanding}} & \multicolumn{4}{c}{\textbf{Pedagogy}} \\
\cmidrule(lr){2-3} \cmidrule(lr){4-6} \cmidrule(lr){7-10}
\textbf{Model} & 
\makecell{Problem \\ solving \\ acc.} & 
\makecell{Socratic \\ questioning \\ BLEU} & 
\makecell{Solution \\ correctness \\ F1} & 
\makecell{Mistake \\ location \\ micro F1} & 
\makecell{Mistake \\ correction \\ acc.} & 
\makecell{scaff. \\ win rate} & 
\makecell{ped.IF \\ win rate} & 
\makecell{scaff. [hard] \\ win rate} & 
\makecell{ped.IF [hard] \\ win rate} \\
\midrule
\multicolumn{10}{c}{\textcolor{violet!70}{\textit{Backbone Model}}} \\
Qwen2.5-7B-Instruct & 0.87 & 0.23 & 0.63 & 0.39 & 0.04 & 0.37 & 0.60 & 0.45 & 0.56 \\
\midrule
UCO w/ Gemini-2.5-Pro & \textbf{0.89} & \textbf{0.24} & \textbf{0.65} & \textbf{0.40} & \textbf{0.06} & \textbf{0.60} & \textbf{0.66} & \textbf{0.63} & \textbf{0.68} \\
UCO w/ DeepSeek-V3.2 & \textbf{0.89} & \textbf{0.24} & \textbf{0.65} & 0.39 & \textbf{0.06} & 0.56 & 0.62 & 0.57 & 0.62 \\
UCO w/ Qwen3-235B-A22B & 0.87 & 0.23 & 0.64 & 0.39 & 0.05 & 0.54 & 0.61 & 0.53 & 0.58 \\
\bottomrule
\end{tabular}
\caption{Comprehensive evaluation on MathTutorBench across various oracle models (500 training samples).}
\label{tab:oracle_math_bench}
\end{table*}

\subsection{\textbf{Impact of Oracle Model Selection}}\label{app:oracle}
To evaluate the sensitivity of UCO to the choice of Oracle models, we conducted comparative experiments using three distinct models: the closed-source Gemini-2.5-Pro~\cite{comanici2025gemini}, the open-source DeepSeek-V3.2~\cite{liu2025deepseek}, and Qwen3-235B-A22B-Instruct \textbf{(Qwen3-235B-A22B)}~\cite{yang2025qwen3}. As shown in Table~\ref{tab:oracle_pedagogy} and Table~\ref{tab:oracle_math_bench}, these models demonstrate highly consistent performance across core dimensions, including the improvement in solve rates, mathematical expertise, and student understanding. 
The variations among the models are marginal and primarily confined to the dimension of pedagogical quality. As detailed in the tables, Gemini-2.5-Pro holds a slight advantage in minimizing solution leakage (16.4\%) and achieving higher overall pedagogical win rates against the backbone. This advantage is likely attributable to a richer integration of educational corpora within its pre-training data. Nevertheless, the UCO, driven by open-source models, not only substantially and comprehensively outperforms the original backbone model across all metrics but also maintains a gap relative to the closed-source frontier model that remains within an acceptable range. Overall, the experimental results verify that UCO exhibits strong robustness to the choice of Oracle models.

\subsection{\textbf{Impact of Rollout on Training Efficiency}}\label{app:rollout} 
\begin{figure}[htbp]
    \centering
    \includegraphics[width=0.48\textwidth]{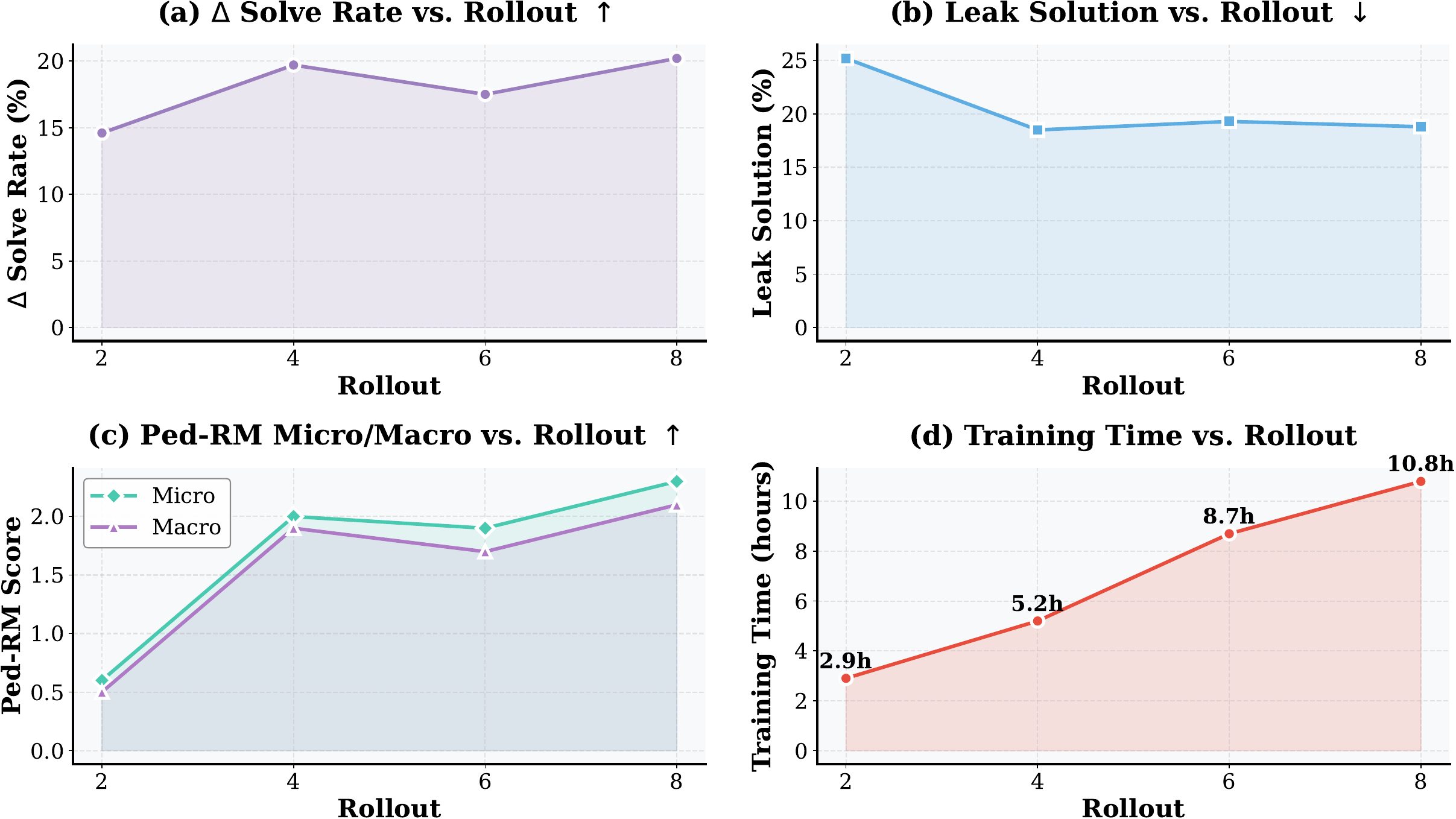}
    \caption{Impact of rollout numbers on model performance and training efficiency.
    }
    \label{fig:your_label}
\end{figure}

We conduct comprehensive ablation studies on rollout numbers using a smaller training set of 100 samples. Figure~\ref{fig:your_label} shows the impact of different rollouts on three key metrics. As rollout numbers increase progressively from 2 to 8, performance improves consistently. Figure~\ref{fig:your_label}(a) shows that $\Delta$Solve Rate increases from $14.6\%$ to $20.2\%$. Figure~\ref{fig:your_label}(b) demonstrates that Leak Solution decreases from $25.2\%$ to $18.8\%$. Figure~\ref{fig:your_label}(c) illustrates that Ped-RM scores also improve steadily, with micro scores rising from $0.6$ to $2.3$ and macro scores from $0.5$ to $2.1$.
However, training cost grows substantially with rollout numbers. Figure~\ref{fig:your_label}(d) shows the detailed training time comparison. With 100 training samples, rollout=2 takes approximately 2.9 hours. Rollout=8 requires 10.8 hours, a 3.7-fold increase. We choose rollout=4 as our default setting, which achieves comparable performance to rollout=8 while reducing training time by $52\%$ (5.2 hours vs 10.8 hours). This provides the optimal trade-off between performance and efficiency.
\begin{table}[htbp]
\centering
\renewcommand{\arraystretch}{1.1}
\scriptsize
\setlength{\tabcolsep}{13pt}
\begin{tabular}{c|c|c|c}
\toprule
\textbf{$\lambda$} & 
\makecell{\textbf{$\Delta$Solve} \\ \textbf{Rate (\%)}} & 
\makecell{\textbf{Leak} \\ \textbf{Solution (\%)}} & 
\makecell{\textbf{Ped-RM} \\ \textbf{micro/macro}} \\
& {\tiny $\uparrow$} & {\tiny $\downarrow$} & {\tiny $\uparrow$} \\
\midrule
0.1 & 26.5 & 16.8 & 3.5/3.6 \\
0.3 & 27.8 & 15.3 & 3.8/4.0 \\
0.5 & \textbf{30.2} & \textbf{12.9} & \textbf{4.6}/\textbf{4.5} \\
0.7 & 29.1 & 13.5 & 4.3/4.2 \\
0.9 & 27.4 & 14.9 & 3.9/3.7 \\
\bottomrule
\end{tabular}
\caption{Ablation study on $\lambda$ balancing potential capacity and semantic quality in progress reward.}
\label{tab:lambda_ablation}
\end{table}

\subsection{\textbf{Impact of Balance Coefficient $\lambda$}} \label{app:lambda}
Table~\ref{tab:lambda_ablation} examines how $\lambda$ balances potential capacity and semantic quality in progress rewards. At low values ($\lambda=0.1$ and $0.3$), semantic quality dominates the reward signal. Rewards depend only on output correctness. Even when the student internally understands correct reasoning, rewards stay low if the output is inaccurate. It prevents the teacher from adjusting teaching strategies to guide the student. This imbalance restricts the model's ability to recognize internal cognitive improvements effectively. It limits improvements in $\Delta$Solve Rate and teaching quality scores.
At high values ($\lambda=0.7$ and $0.9$), potential capacity dominates the reward signal. Rewards mainly depend on increases in internal confidence and ignore output accuracy. To quickly boost student confidence, the teacher may directly provide answer-related information rather than guide autonomous reasoning. It increases answer leakage rates while decreasing $\Delta$Solve Rate and Ped-RM scores.
At $\lambda=0.5$, performance reaches the optimum. The potential capacity component measures student confidence in correct reasoning. The semantic quality component verifies output correctness. Together, they enable the teacher to optimize both reasoning capability and expression quality. It achieves the best performance across all metrics.

\subsection{\textbf{Impact of Temperature Coefficient $\alpha$}}\label{app:alpha}
Table~\ref{tab:alpha_ablation} examines how the temperature coefficient $\alpha$ affects reward signal strength in potential ability scoring. At lower values ($\alpha=1.0, 1.5$), the tanh function produces weak reward signals. Consequently, these signals cannot effectively distinguish confidence levels in student responses. This insufficient discrimination prevents the teacher model from identifying meaningful cognitive progress. 
\begin{table}[htbp]
\centering
\renewcommand{\arraystretch}{1.1}
\scriptsize
\setlength{\tabcolsep}{13pt}
\begin{tabular}{c|c|c|c}
\toprule
\textbf{$\alpha$} & 
\makecell{\textbf{$\Delta$Solve} \\ \textbf{Rate (\%)}} & 
\makecell{\textbf{Leak} \\ \textbf{Solution (\%)}} & 
\makecell{\textbf{Ped-RM} \\ \textbf{micro/macro}} \\
& {\tiny $\uparrow$} & {\tiny $\downarrow$} & {\tiny $\uparrow$} \\
\midrule
1.0 & 27.4 & 15.1 & 3.8/4.0 \\
1.5 & 28.9 & 13.8 & 4.1/4.2 \\
2.0 & \textbf{30.2} & \textbf{12.9} & \textbf{4.5}/\textbf{4.4} \\
2.5 & 28.6 & 14.3 & 4.2/4.0 \\
3.0 & 27.1 & 16.1 & 3.3/2.8 \\
\bottomrule
\end{tabular}
\caption{Ablation study on temperature $\alpha$ controlling reward sensitivity in potential capacity score.}
\label{tab:alpha_ablation}
\end{table}
As a result, without clear signals, the model cannot guide students toward correct reasoning paths, leading to poor teaching performance across all evaluation metrics.
At higher values ($\alpha=2.5, 3.0$), reward signals become overly sensitive to small confidence fluctuations. They amplify minor probability changes into large reward differences. This excessive sensitivity introduces noise into the training, which disrupts stability and leads to inconsistent teaching behavior. Such instability compromises the model's overall performance.
At the optimal value ($\alpha=2.0$), the temperature coefficient achieves an ideal balance. Reward signals maintain sufficient discriminative strength while preserving stability against noise amplification. This balance enables the teacher model to reliably identify genuine cognitive improvements and deliver consistent, effective teaching strategies.

\subsection{\textbf{Impact of Semantic Quality Threshold $\delta$}}\label{app:delta}
Table~\ref{tab:delta_ablation} examines how the bias term $\delta$ affects output filtering in semantic quality scoring.
At lower values ($\delta=0.5, 0.6$), the threshold is too lenient. Consequently, the model cannot effectively distinguish genuine understanding from shallow pattern matching. This causes the reward function to inappropriately reward superficial matches lacking true reasoning, resulting in teaching strategies that lack precision and effectiveness.
At higher values ($\delta=0.8, 0.9$), the threshold becomes overly strict. Consequently, many responses demonstrating valid understanding are incorrectly filtered out. This causes the reward signal to inadequately capture the actual progression in students' cognitive development. As a result, the teacher model tends to provide excessive scaffolding, which undermines students' autonomous reasoning capacity and increases answer leakage risk.
At the optimal value $\delta=0.7$, the threshold achieves an ideal balance. 
Unlike lower values that allow noisy signals to mislead the optimization process, or higher values that discard valuable learning signals, this optimal setting enables the model to accurately filter ambiguous outputs while reliably identifying genuine 
\begin{table}[htbp]
\centering
\renewcommand{\arraystretch}{1.1}
\scriptsize
\setlength{\tabcolsep}{13pt}
\begin{tabular}{c|c|c|c}
\toprule
\textbf{$\delta$} & 
\makecell{\textbf{$\Delta$Solve} \\ \textbf{Rate (\%)}} & 
\makecell{\textbf{Leak} \\ \textbf{Solution (\%)}} & 
\makecell{\textbf{Ped-RM} \\ \textbf{micro/macro}} \\
& {\tiny $\uparrow$} & {\tiny $\downarrow$} & {\tiny $\uparrow$} \\
\midrule
0.5 & 27.2 & 16.1 & 3.7/3.9 \\
0.6 & 28.6 & 14.5 & 4.0/4.2 \\
0.7 & \textbf{30.2} & \textbf{12.9} & \textbf{4.6}/\textbf{4.5} \\
0.8 & 29.3 & 13.7 & 4.2/4.3 \\
0.9 & 26.8 & 17.3 & 3.6/3.8 \\
\bottomrule
\end{tabular}
\caption{Ablation study on semantic threshold $\delta$ for filtering ambiguous outputs in semantic quality score.}
\label{tab:delta_ablation}
\end{table}
\begin{table}[htbp]
\centering
\renewcommand{\arraystretch}{1.1}
\scriptsize
\setlength{\tabcolsep}{13pt}
\begin{tabular}{c|c|c|c}
\toprule
\textbf{$c$} & 
\makecell{\textbf{$\Delta$Solve} \\ \textbf{Rate (\%)}} & 
\makecell{\textbf{Leak} \\ \textbf{Solution (\%)}} & 
\makecell{\textbf{Ped-RM} \\ \textbf{micro/macro}} \\
& {\tiny $\uparrow$} & {\tiny $\downarrow$} & {\tiny $\uparrow$} \\
\midrule
0.1 & 27.6 & 15.8 & 3.8/3.9 \\
0.2 & \textbf{30.2} & \textbf{12.9} & \textbf{4.6}/\textbf{4.5} \\
0.3 & 28.7 & 14.1 & 4.2/4.1 \\
0.4 & 27.3 & 15.6 & 3.9/4.0 \\
0.5 & 26.1 & 17.2 & 3.5/3.7 \\
\bottomrule
\end{tabular}
\caption{Ablation study on deviation coefficient $c$ regulating scaffolding precision in scaffolding reward.}
\label{tab:c_ablation}
\end{table}
understanding. 
Consequently, the reward function precisely evaluates students' cognitive states, and performance is optimized across all metrics.

\subsection{\textbf{Impact of Deviation Coefficient $c$.}}\label{app:deviation}
Table~\ref{tab:c_ablation} examines how the deviation penalty coefficient $c$ regulates scaffolding precision. This coefficient penalizes teaching strategies that deviate from students' Zone of Proximal Development (ZPD).
At lower values ($c=0.1$), the penalty strength is insufficient. The model receives weak feedback signals for scaffolding deviations. The teacher model struggles to effectively adjust teaching difficulty to appropriate levels. All evaluation metrics show suboptimal performance.
At higher values ($c=0.3, 0.4, 0.5$), penalty strength becomes overly strict. The model's teaching strategies are over-constrained. The teacher model's ability to respond to individual student needs declines accordingly. Limited teaching flexibility leads to overall performance degradation.
At the optimal value ($c=0.2$), penalty strength achieves an ideal balance. The model receives sufficient incentive to maintain scaffolding precisely within the ZPD range. All evaluation metrics achieve optimal performance under this configuration.

\begin{table}[htbp]
\centering
\renewcommand{\arraystretch}{1.1}
\scriptsize
\setlength{\tabcolsep}{6pt}
\begin{tabular}{l|c|c}
\toprule
\textbf{Model} & \textbf{GSM8K} & \textbf{MATH500} \\
\midrule
\multicolumn{3}{c}{\textcolor{violet!70}{\textit{Backbone Model}}} \\
Qwen2.5-7B-Instruct~\cite{team2024qwen2} & 86.8 & 75.4 \\
\midrule
\multicolumn{3}{c}{\textcolor{violet!70}{\textit{Education-specific Models}}} \\
Qwen2.5-7B-SFT~\cite{macina2023mathdial} & 79.5 & 66.0 \\
Qwen2.5-7B-MDPO~\cite{xiongbuilding} & \textbf{87.0} & \textbf{76.4} \\
Qwen2.5-7B-RL~\cite{dinucu2025problem} & 86.1 & 73.6 \\
\midrule
\textbf{UCO} & 85.9 & 75.0 \\
\bottomrule
\end{tabular}
\caption{Evaluation of general mathematical reasoning benchmarks GSM8K and MATH500.}
\label{tab:math_results}
\end{table}

\subsection{\textbf{General Mathematical Reasoning Ability Preservation}}\label{app:general_math} Table~\ref{tab:math_results} shows UCO's performance on general math reasoning benchmarks. UCO achieves $85.9\%$ and $75.0\%$ accuracy on GSM8K and MATH500. 
This represents only $0.9$ and $0.4$ percentage point drops from the backbone model. Other education-specific models show larger performance declines. Qwen2.5-7B-SFT drops by $7.3$ and $9.4$ percentage points. Qwen2.5-7B-RL drops by $0.7$ and $1.8$ percentage points. Qwen2.5-7B-MDPO performs best on general ability. However, it shows very limited improvement in teaching quality. This demonstrates that UCO achieves the best trade-off between teaching specialization and general ability.

\subsection{\textbf{Human Evaluation.}}\label{app:human_eval}
To further evaluate the effectiveness of UCO, we invited three experienced teachers to conduct a human assessment. Specifically, we randomly sampled 100 problems from the BigMath test set. Following the experimental setup of PedagogicalRL~\cite{dinucu2025problem}, various models acted as teachers to interact with a unified student model (Meta-Llama-3.1-8B-Instruct) to generate complete multi-turn teacher-student dialogues. The three teachers assessed the multi-turn teacher-student dialogues generated by different models against those produced by GPT-4o to determine the Win Rate. This evaluation is from five dimensions proposed by SocraticLM~\cite{liu2024socraticlm}, namely Overall Quality, Incorrect Answer Recognition Accuracy (IARA), Correct Answer Recognition Accuracy (CARA), Successful Explanation Rate (SER), and Successful Rejection Rate (SRR).

As shown in Table~\ref{tab:human_eval_results}, UCO achieved a win rate of 0.69 against GPT-4o, significantly outperforming all baseline methods. Results demonstrate that UCO can accurately perceive student states and dynamically adjust teaching strategies. Compared to GPT-4o, which tends to provide solutions directly in the later stages of a conversation, UCO exhibits superior scaffolding and guidance capabilities.

\begin{table}[htbp]
\centering
\renewcommand{\arraystretch}{1.1}
\scriptsize
\setlength{\tabcolsep}{14pt}
\begin{tabular}{lc}
\toprule
\textbf{Teacher Model} & \textbf{Win Rate (vs. GPT-4o)} \\
\midrule
\multicolumn{2}{c}{\textcolor{violet!70}{\textit{Backbone Model}}} \\
Qwen2.5-7B-Instruct & 0.24 \\
\midrule
GPT-4o & 0.50 \\
SocraticLM & 0.57 \\
Qwen2.5-7B-RL & 0.61 \\
\midrule
\textbf{UCO (Ours)} & \textbf{0.69} \\
\bottomrule
\end{tabular}
\caption{Human evaluation results on BigMath.}
\label{tab:human_eval_results}
\end{table}

\subsection{Case Study}\label{app:case_study}
To verify the effectiveness of our proposed method, we conduct a detailed qualitative case study. We compare our model with GPT-4o-2024-11-20~\cite{hurst2024gpt}, both serving as teacher models. The student model uses Meta-Llama-3.1-8B-Instruct. We carefully select two representative problems from the BigMath~\cite{albalak2025big} test set. The maximum dialogue turns are set to 16. The teacher model outputs \texttt{<end\_of\_conversation>} when it determines the conversation should end. Following prior research~\cite{dinucu2025problem}, the specific prompt designs for teacher and student models are shown in Table~\ref{tab:prompts}.
\begin{table}[h]
\centering
\scriptsize
\begin{tabular}{p{0.42\linewidth}|p{0.42\linewidth}}
\toprule
\textbf{UCO} & \textbf{GPT-4o} \\
\midrule
\multicolumn{2}{p{0.9\linewidth}}{\textit{Problem: Gina can paint 6 rose cups/hour and 7 lily cups/hour. Her store receives an order for 6 rose cups and 14 lily cups. If she gets \$90 total, what is her hourly pay?}} \\
\midrule
\textbf{T:} To calculate a "dollars per hour" rate, what two core pieces of information do we need? & 
\textbf{T:} Let's solve this. What do you think we should calculate first? \\
\textbf{S:} The total dollars she made... and the total hours she worked? & 
\textbf{S:} Um... maybe how long it takes to finish the order? \\
\textbf{T:} Exactly. Total pay is given. So the real task is finding total hours. The order has two parts. How would you approach that? & 
\textbf{T:} Good thinking. To get total time, we look at two cup types separately. For 6 rose cups at 6/hour, that takes 1 hour. Clear? \\
\textbf{S:} I'd figure out time for roses, then lilies, and add them. & 
\textbf{S:} Clear, that's simple. \\
\textbf{T:} That's a solid plan. Go ahead, what's the time for roses? & 
\textbf{T:} The other part is 14 lily cups. She paints 7/hour. Can you calculate how long that takes? \\
\textbf{S:} She paints 6 at 6/hour... so 1 hour. & 
\textbf{S:} 14 divided by 7 is 2 hours. \\
\textbf{T:} Right. And the lilies? & 
\textbf{T:} Right. So total time is 1 + 2 = 3 hours. Now we know total time and pay, what's the final step? \\
\textbf{S:} 14 cups at 7/hour... 14÷7 = 2 hours. & 
\textbf{S:} Um... use \$90 divided by 3 hours? \\
\textbf{T:} Correct. Now you have everything. What's the final step? & 
\textbf{T:} Yes, that gives her hourly rate. \\
\textbf{S:} Total time is 3 hours. So \$90÷3 = \$30/hour. & 
\textbf{S:} The result is \$30/hour. \\
\textbf{T:} You've walked through the solution logically. \texttt{<end\_of\_conversation>} & 
\textbf{T:} Right, the final answer is \$30. \texttt{<end\_of\_conversation>} \\
\bottomrule
\end{tabular}
\caption{Case One Dialogue Comparison}
\label{tab:case1_dialogue}
\end{table}
We present two dialogue cases and conduct an in-depth comprehensive analysis to reveal the core advantages of our method:

For case one, this problem involves basic rate calculation. Table~\ref{tab:case1_dialogue} shows the complete dialogue process of both models.

Our method demonstrates stronger adaptive guidance ability. First, our model activates students' metacognitive thinking through open-ended questions (``What two core pieces of information do we need?''). This heuristic questioning forces students to identify the problem's essential structure. Students do not passively accept solution steps. In contrast, GPT-4o directly states the solution direction (``we look at two cup types separately''). It deprives students of independent thinking opportunities.
Second, our method exhibits more precise progressive scaffolding. When the student proposes a complete plan, the teacher model only gives affirmation and encourages execution (``That's a solid plan. Go ahead''). It transfers complete computational agency to the student. GPT-4o directly provides intermediate calculation results (``that takes 1 hour''). It interferes with students' autonomous problem-solving process. Such over-guidance diminishes students' cognitive engagement and independent reasoning.
Finally, our method emphasizes students' thinking process when ending the dialogue (``You've walked through the solution logically''). 
\begin{table}[h]
\centering
\scriptsize
\begin{tabular}{p{0.42\linewidth}|p{0.42\linewidth}}
\toprule
\textbf{UCO} & \textbf{GPT-4o} \\
\midrule
\multicolumn{2}{p{0.9\linewidth}}{\textit{Problem: $f(x)$ on $\mathbb{R}$ is even, satisfies $f(x+1)=-f(x)$. When $x \in [-1,0)$, $f(x)=(\frac{1}{2})^x$. Find $f(\log_{2}8)$.}} \\
\midrule
\textbf{T:} This gives three pieces of info about $f(x)$. First, it's "even." Recall its definition? & 
\textbf{T:} This looks complex, but the key is using given properties. We need $f(\log_{2}8)$, which is $f(3)$. \\
\textbf{S:} I think it's $f(-x) = f(x)$. & 
\textbf{S:} But the formula's domain is $[-1, 0)$, and 3 isn't in it. \\
\textbf{T:} Correct. Second, $f(x+1) = -f(x)$. This tells what happens when input increases by 1. What about increasing by 2? & 
\textbf{T:} Correct. We use properties to transform $f(3)$. Which helps? \\
\textbf{S:} Hmm... $f(x+2) = -f(x+1)$. Since $f(x+1) = -f(x)$, then $f(x+2) = f(x)$. & 
\textbf{S:} I'm not sure... the periodic one? \\
\textbf{T:} Your reasoning is correct. What does this property tell us? & 
\textbf{T:} Yes. We deduce $f(x+2) = f(x)$, so period is 2. Using this, $f(3) = f(1)$. \\
\textbf{S:} The function is periodic with period 2. & 
\textbf{S:} Now we need $f(1)$. \\
\textbf{T:} Very good. Now what we need: $f(\log_{2}8)$. What is $\log_{2}8$? & 
\textbf{T:} Let's use $f(x+1) = -f(x)$. When $x=0$, this becomes $f(1) = -f(0)$. \\
\textbf{S:} 2 to power 3 is 8, so it's 3. We need $f(3)$. & 
\textbf{S:} Oh, so we just need $f(0)$. \\
\textbf{T:} Right. How can we use the period property? & 
\textbf{T:} Right. We use $f(x+1)=-f(x)$ again. When $x=-1$, we get $f(0)=-f(-1)$. \\
\textbf{S:} $f(3) = f(1)$. & 
\textbf{S:} So $f(-1) = (1/2)^{-1} = 2$. \\
\textbf{T:} Good. Now the problem becomes finding $f(1)$. Which condition helps? & 
\textbf{T:} Correct. So $f(0) = -2$. \\
\textbf{S:} $f(x+1) = -f(x)$? & 
\textbf{S:} Then $f(1) = -f(0) = 2$. \\
\textbf{T:} Yes, how would you use it? & 
\textbf{T:} Exactly. Since $f(3) = f(1)$, what's the final answer? \\
\textbf{S:} If I let $x=0$, I get $f(1) = -f(0)$. & 
\textbf{S:} The final answer is 2. \\
\textbf{T:} Excellent. How can we find $f(0)$? & 
\textbf{T:} Correct. The answer is 2. \texttt{<end\_of\_conversation>} \\
\textbf{S:} If I let $x=-1$, I get $f(0) = -f(-1)$. & \\
\textbf{T:} Perfect! What is $f(-1)$? & \\
\textbf{S:} $f(-1) = (1/2)^{-1} = 2$. & \\
\textbf{T:} Correct. So what is $f(0)$? & \\
\textbf{S:} $f(0) = -2$. & \\
\textbf{T:} Can you now connect all results? & \\
\textbf{S:} Since $f(0)=-2$, then $f(1)=2$. And $f(3)=2$. & \\
\textbf{T:} Exactly right! You successfully applied multiple function properties to complete the derivation. \texttt{<end\_of\_conversation>} & \\
\bottomrule
\end{tabular}
\caption{Case Two Dialogue Comparison}
\label{tab:case2_dialogue}
\end{table}
It strengthens students' metacognitive reflection on solution strategies. GPT-4o only confirms answer correctness. It ignores the summary of the thinking process.

For case two, this problem examines students' comprehensive understanding of periodicity and parity. Table~\ref{tab:case2_dialogue} shows strategies of both models when handling abstract mathematical concepts.

When handling complex abstract problems, our method shows finer cognitive scaffolding construction ability. First, our model verifies students' basic concept understanding (``Recall its definition?''). It guides students to independently derive periodicity (``What about increasing by 2?''). It promotes students to abstract derivation results into mathematical properties (``What does this property tell us?''). 
\begin{table}[h]
\centering
\footnotesize
\begin{tabular}{p{0.92\linewidth}}
\toprule
\textbf{Teacher Prompt} \\
\midrule
You are tasked with being a teacher and helping a student with a math problem. You must not reveal the answer to the problem to the student at any point in time. Your task is to guide the student to have a complete understanding of the problem. Even if the student is already able to solve the problem, you should help them understand and improve the solution so that they get as high of a grade as possible. If possible, do not respond with overly long responses to the student. You can end a conversation by writing \texttt{<end\_of\_conversation>}, please try to end conversations as soon as they are finished instead of prolonging them if not needed. But do not end them prematurely either. Here is the math problem: \{\{ problem \}\}. \\
\midrule
\textbf{Student Prompt} \\
\midrule
You will act as a student in a conversation with a teacher in training. You will need to act as much like a student as possible. If possible do not respond with overly long messages. The conversation with the teacher will be about this math problem: \{\{ problem \}\}. You may or may not know how to solve it already, let the teacher guide you to the correct understanding. You will be tested at the end and scored thus it is best if you collaborate with the teacher as it has more experience in math than you. \\
\bottomrule
\end{tabular}
\caption{Teacher and Student Prompts}
\label{tab:prompts}
\end{table}
This structured guidance cultivates students' mathematical reasoning ability and concept transfer ability.
GPT-4o skips the concept verification stage. It directly tells students ``We deduce $f(x+2) = f(x)$'' and ``period is 2''. While expedient, this approach limits students' opportunities for independent mathematical discovery. 
It hinders the formation of a deep understanding.
Second, our method requires active student participation at each reasoning node (``Which condition helps?'' ``How would you use it?''). This continuous cognitive activation ensures students remain in the leading position of problem-solving. GPT-4o tends to provide explicit procedural directions (``Let's use $f(x+1) = -f(x)$''). Students only need to execute the given steps.
Finally, our method strengthens students' overall understanding of multi-step reasoning through comprehensive questions (``Can you now connect all results?''). It facilitates the transfer of the application to similar problems.

Through comparative analysis of these two cases, our method demonstrates stronger adaptive teaching ability. It reflects the effectiveness of the proposed reward function.

\subsection{\textbf{Efficiency Analysis.}} \label{app:efficiency}
We analyze the efficiency of UCO from three dimensions: training cost, inference latency, and scalability. During training, UCO requires approximately 112 hours on 8 NVIDIA H200 GPUs to complete the full training process, compared to 48 hours for Qwen2.5-7B-RL, with the additional overhead primarily stemming from interactions with the oracle model during each dialogue turn. However, this cost is incurred during the training phase. At inference time, UCO is essentially a standard Qwen2.5-7B-Instruct model that no longer relies on the oracle model or the student model, so its inference latency is fully consistent with baselines of the same scale, with single-turn teacher response generation taking approximately 0.3–0.5 seconds on a single H200 GPU, meeting the interaction requirements of real-time tutoring. Furthermore, the 7B parameter size allows UCO to perform efficient inference on a single GPU, enabling it to provide tutoring services at low hardware cost. 

\subsection{Evaluation on Additional Benchmarks}
\label{app:mathdial}
To validate the effectiveness of UCO across broader evaluation benchmarks, we adopted the dual-dimensional framework from Tutorbot-DPO~\cite{scarlatos2025training} on the MathDial test set, assessing models on \textit{Student Outcomes} and \textit{Pedagogical Principles}. 
As shown in Table~\ref{tab:baseline_comparison}, UCO surpasses Tutorbot-DPO in the overall pedagogical score, with significant advantages in advancing progress, hinting, and encouraging students. Conversely, it scores slightly lower on student outcomes and direct error correction. This performance difference reflects a fundamental divergence in teaching strategies: Tutorbot-DPO optimizes for single-turn correctness, causing the model to directly point out errors and provide explicit answers. 
In contrast, UCO leverages online reinforcement learning to provide progressive hints. Although the multi-turn elicitation process slightly decreases the immediate correctness metric, it enables the model to dynamically adjust its scaffolding based on the student's state. This deliberate trade-off, which prioritizes long-term cognitive engagement over short-term immediate metrics, better aligns with established educational principles.

\begin{table}[htbp]
\centering
\renewcommand{\arraystretch}{1.1}
\scriptsize
\setlength{\tabcolsep}{2.2pt} 
\begin{tabular}{l | c | cccccc | c}
\toprule
& \textbf{Student} & \multicolumn{6}{c|}{\textbf{Pedagogical Principles}} & \textbf{Overall} \\
\cmidrule(lr){3-8}
\textbf{Method} & 
\makecell{\textbf{Outcomes} \\ \textbf{$\uparrow$}} & 
\makecell{\textbf{Acc.} \\ \textbf{$\uparrow$}} & 
\makecell{\textbf{Prog.} \\ \textbf{$\uparrow$}} & 
\makecell{\textbf{Err.} \\ \textbf{$\uparrow$}} & 
\makecell{\textbf{Hint} \\ \textbf{$\uparrow$}} & 
\makecell{\textbf{Wth.} \\ \textbf{$\uparrow$}} & 
\makecell{\textbf{Enc.} \\ \textbf{$\uparrow$}} & 
\makecell{\textbf{Score} \\ \textbf{$\uparrow$}} \\
\midrule
Tutorbot-DPO & \textbf{0.65} & \textbf{0.97} & 0.96 & \textbf{0.92} & 0.96 & \textbf{0.99} & 0.92 & 9.37 \\
\midrule
UCO & 0.63 & 0.95 & \textbf{0.98} & 0.89 & \textbf{0.99} & \textbf{0.99} & \textbf{0.98} & \textbf{9.48} \\
\bottomrule
\end{tabular}
\caption{Evaluation of student outcomes and pedagogical principles on the MathDial benchmark. Pedagogical metrics include Accuracy (\textbf{Acc.}), Progress (\textbf{Prog.}), Error identification (\textbf{Err.}), Strategic Hinting (\textbf{Hint}), Withholding the answer (\textbf{Wth.}), and Encouraging (\textbf{Enc.}).}
\label{tab:baseline_comparison}
\end{table}

\end{document}